\documentclass[]{interact}

\usepackage[numbers,sort&compress]{natbib}% Citation support using natbib.sty
\bibpunct[, ]{[}{]}{,}{n}{,}{,}% Citation support using natbib.sty
\makeatletter% @ becomes a letter
\def\NAT@def@citea{\def\@citea{\NAT@separator}}% Suppress spaces between citations using natbib.sty
\makeatother% @ becomes a symbol again

\usepackage{graphicx}% for atom
\usepackage{algpseudocode,algorithm}
\usepackage{amsmath,amssymb,amsfonts}
\usepackage{subcaption}

\theoremstyle{plain}% Theorem-like structures provided by amsthm.sty

\theoremstyle{definition}

\theoremstyle{remark}

\begin{document}

\jvol{00} \jnum{00} \jyear{2013} \jmonth{January}

\articletype{Full Papers}

\title{
Design of Restricted Normalizing Flow towards \\Arbitrary Stochastic Policy with Computational Efficiency
}

\author{
\name{
Taisuke Kobayashi\textsuperscript{a}\thanks{CONTACT T. Kobayashi. Email: kobayashi@nii.ac.jp}
 and Takumi Aotani\textsuperscript{b}
}
\affil{
\textsuperscript{a}Principles of Informatics Research Division, National Institute of Informatics, Tokyo, Japan; and School of Multidisciplinary Sciences, Department of Informatics, The Graduate University for Advanced Studies (SOKENDAI), Kanagawa, Japan
\\
\textsuperscript{b}Department of Mechanical Engineering Informatics, School of Science and Technology, Meiji University, Japan
}
}

\maketitle

%%%%%%%%%%%%%%%%%%%%%%%%%%%%%%%%%%%%%%%%%%%%%%%%%%%%%%%%%%%%%%%%%%%%%%%%%%%%%%%%
\begin{abstract}

This paper proposes a new design method for a stochastic control policy using a normalizing flow (NF).
In reinforcement learning (RL), the policy is usually modeled as a distribution model with trainable parameters.
When this parameterization has less expressiveness, it would fail to acquiring the optimal policy.
A mixture model has capability of a universal approximation, but it with too much redundancy increases the computational cost, which can become a bottleneck when considering the use of real-time robot control.
As another approach, NF, which is with additional parameters for invertible transformation from a simple stochastic model as a base, is expected to exert high expressiveness and lower computational cost.
However, NF cannot compute its mean analytically due to complexity of the invertible transformation, and it lacks reliability because it retains stochastic behaviors after deployment for robot controller.
This paper therefore designs a restricted NF (RNF) that achieves an analytic mean by appropriately restricting the invertible transformation.
In addition, the expressiveness impaired by this restriction is regained using bimodal student-t distribution as its base, so-called Bit-RNF.
In RL benchmarks, Bit-RNF policy outperformed the previous models.
Finally, a real robot experiment demonstrated the applicability of Bit-RNF policy to real world.

\end{abstract}


%%%%%%%%%%%%%%%%%%%%%%%%%%%%%%%%%%%%%%%%%%%%%%%%%%%%%%%%%%%%%%%%%%%%%%%%%%%%%%%%
\section{Introduction}

%%% robot learning
The control target in robotic applications is beginning to contain unknown or highly nonlinear models:
e.g.
human at physical human-robot interaction~\cite{modares2015optimized,kobayashi2021whole};
rough terrain at disaster sites~\cite{kobayashi2015selection,delmerico2019current};
and general objects at robot manipulation~\cite{tsurumine2019deep,kroemer2021review}.
Classic model-based control would be intractable to be applied into these tasks due to limitation of mathematical modeling in advance.
To resolve them, machine learning based control, e.g. reinforcement learning (RL)~\cite{sutton2018reinforcement} and imitation learning~\cite{hussein2017imitation}, has recently been studied with the expectation that it can find the rules necessary for control from experienced data with no or minimal domain knowledge.
In particular, deep neural networks (DNNs)~\cite{lecun2015deep} have yield high expressiveness for function approximation in such a control method (e.g. policy and value functions in RL).

%%% policy model
Although it is true that DNNs provide high expressiveness in function approximation, what function to be approximated is more important from the modeling point of view.
In RL, the controller is modeled as a probability distribution, and the parameters contained in the distribution model are approximated by DNNs.
For real-valued controls such as joint torques, normal distribution is widely employed, and its location and scale parameters are approximated.
However, this modeling is an operation that extracts a subset of probability distributions, and if the extracted subset does not contain the optimal policy, performance degradation will occur.
Alternatively, DNNs are typically trained using stochastic gradient descent method~\cite{robbins1951stochastic,kingma2014adam,ilboudo2022adaterm}, hence it is easier to get trapped in one of local solutions if the modeling causes a lot of them.
In addition, excessive modeling leads to an increase in computational cost and may become a bottleneck of the control period in real-time robot control.
Although natural gradient methods have been proposed to counteract such local solutions~\cite{kakade2001natural,peters2008natural}, the computational cost explodes when used with function approximations with many parameters such as DNNs.

%%% conventional work
For these open issues, several probability distribution models have been developed to model the policy in previous studies.
As an extension of normal distribution, student-t distribution has been employed~\cite{kobayashi2019student}.
It can adjust its heavy tail by a parameter representing degrees of freedom, and results in achieving the capability to find the global optimal solution with high exploration performance and conservative learning.
Modeling with beta distribution has also been proposed in consideration of bounded real-valued control~\cite{chou2017improving}.
This can eliminate the bias caused by forcing normal distribution to be bounded in ad-hoc implementation, although this problem may be mitigated by an invertible map between the bounded and real-valued spaces.
Although a method have been proposed to define the policy function implicitly without explicitly providing such a probability distribution model~\cite{haarnoja2017reinforcement}, this direction would not be practical because even sampling actions from the implicit policy requires approximations.

Mixture distribution, e.g. Gaussian mixture model (GMM), is widely known as the modeling that can represent more complex probability distributions~\cite{park1991universal,park1993approximation}.
In particular, its multimodality has been shown to be useful for problems that can have multiple solutions~\cite{baram2021maximum,sasaki2021variational}.
Theoretically, having an infinite number of components can guarantee a universal approximation, but for implementation, it must be limited to a finite number.
The number of components is difficult to be designed:
if the number is too small, the required expressiveness cannot be obtained;
and if it is too large, the system becomes redundant, which increases computational cost.

%%% normalizing flow
Normalizing flow (NF) has been developed actively as a theoretically computationally efficient methodology while possessing high expressiveness~\cite{papamakarios2021normalizing}.
NF has a relatively simple probability distribution model as its base, and connects it to trainable invertible transformation(s) to represent complex probability distributions.
A variety of computationally efficient invertible transforms have been proposed, and their expressiveness has been shown theoretically to be high~\cite{teshima2020coupling,kong2020expressive}.
Indeed, there are some examples of applying NF to the RL policy, and they reported that NF improves the learning performance~\cite{ward2019improving,mazoure2020leveraging,ma2020normalizing}.
In addition, if we pay attention to the fact that two distributions connected by the invertible transformation are retained, we can expect applications in residual learning, multi-task learning, and style transformation, where common objectives across several tasks are learned in the base and each objective is achieved after the transformation~\cite{gambardella2019transflow,codevilla2018end,abdal2021styleflow}.

%%% no-closed form
This study focuses on such NF capability to model the policy, but one issue arises here.
Namely, the statistics of the probability distribution modeled by NF cannot be given analytically because the complexity of the invertible transformation prevents analytical integral calculations.
As a result, the necessary information for the controller, such as the mean behavior and its accuracy (or uncertainty), cannot be obtained, and the controller lacks reliability in practical use.
In particular, the lack of an analytic mean leaves sampling-based stochastic behavior even when the trained policy is deployed for operating the robot system as inference mode.
During the learning process, this sampling accelerates the exploration, but it is not conducive to producing stable performance.
Note that we have confirmed that the sampling-based control is prone to worst-case scenarios than the case with the analytic mean of the trained policy (see Section~\ref{subsec:unstable_control}).

%Figure
\begin{figure*}[tb]
    \centering
    \includegraphics[keepaspectratio=true,width=0.96\linewidth]{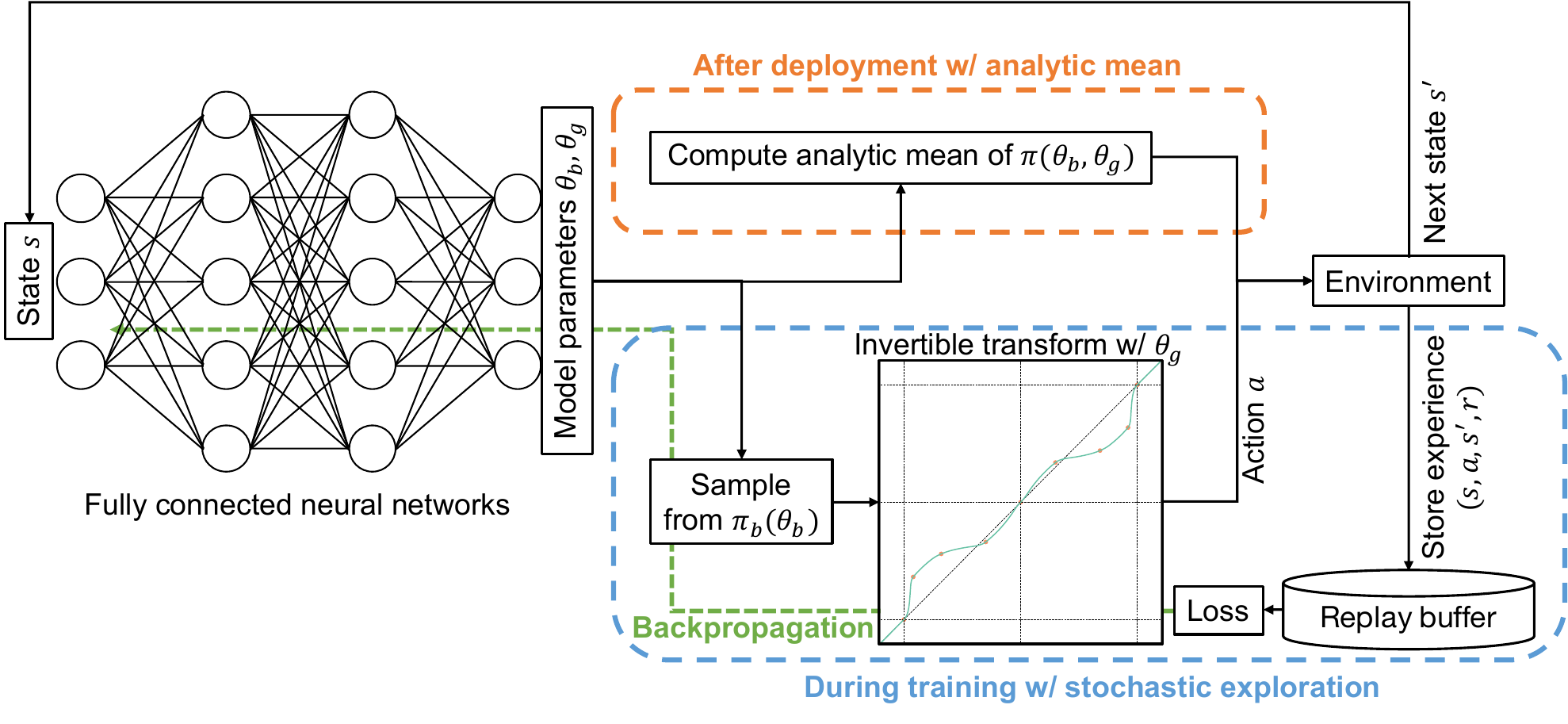}
    \caption{Policy architecture with restricted normalizing flow (RNF):
        during training, RNF provides efficient exploration using highly expressive probability distributions;
        after the deployment of the trained policy, reliable control can be achieved by using the analytic mean of RNF.
    }
    \label{fig:architecture}
\end{figure*}

%%% proposal
Therefore, in this paper, we derive a restricted normalizing flow (RNF) by revealing the constraints that make it possible to compute the analytic mean, which can be utilized after deployment (see Fig.~\ref{fig:architecture}).
We mathematically show that RNF can be achieved by i) separating the mean and the stochastic component of the symmetric probability distribution and ii) applying invertible transformation(s) by odd function(s) only to the probability component.
Among the conventional invertible transformations, linear rational spline (LRS)~\cite{dolatabadi2020invertible} and real-valued non-volume preserving (RealNVP)~\cite{dinh2016density} are restricted as examples.

%%% extension
However, the constraints of the modeling inevitably undermines the expressiveness, and RNF actually has inexpressible probability distributions.
In particular, the proposed constraints make the distribution shape symmetrical.
We need implementation techniques that recover this loss of expressiveness while minimizing the increase in computational cost.
To this end, bimodal student-t (Bit) distribution is employed as the base for RNF.
In this way, we propose a so-called Bit-RNF, which regains the capability to adjust the heavy tail and to represent asymmetric distributions.

%%% result
In order to statistically evaluate the performance of the policy with Bit-RNF, we first compare the conventional and proposed models by numerical simulation benchmarks~\cite{brockman2016openai,coumans2016pybullet}.
Compared with the conventional models, we show that Bit-RNF can achieve consistently high performance in all benchmarks.
In addition, we also demonstrate ball manipulation on a ball-plate system constructed by a delta robot with variable stiffness actuators (VSAs)~\cite{catalano2011vsa}.
Bit-RNF is capable of acquiring this task, while normal distribution fails and GMM does not satisfy the specified control period.

%%%%%%%%%%%%%%%%%%%%%%%%%%%%%%%%%%%%%%%%%%%%%%%%%%%%%%%%%%%%%%%%%%%%%%%%%%%%%%%%
\section{Preliminaries}

%%%%%%%%%%%%%%%%%%%%%%
\subsection{Normalizing Flow}

%Figure
\begin{figure*}[tb]
    \centering
    \includegraphics[keepaspectratio=true,width=0.96\linewidth]{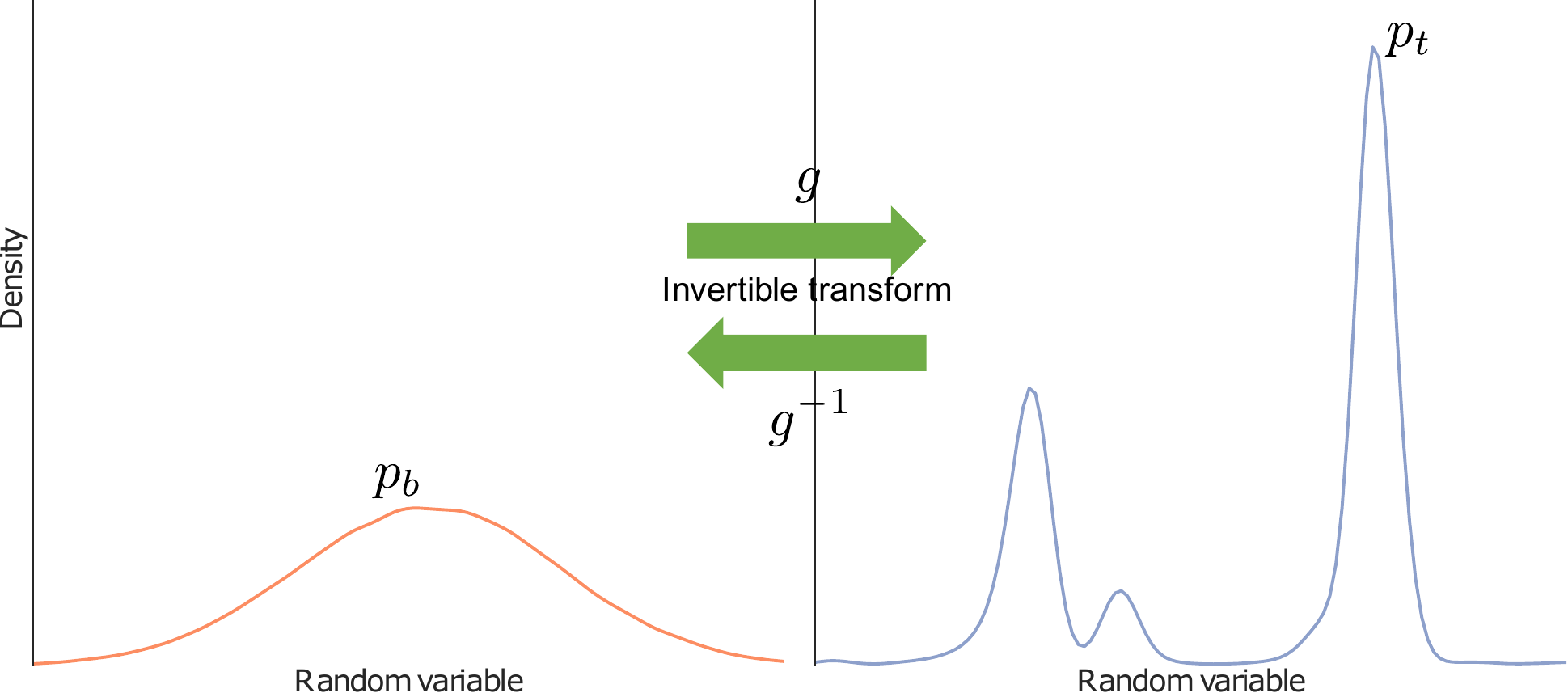}
    \caption{Normalizing flow:
        a simple distribution is set to be a base;
        using invertible transformation(s) $g$, sampled variables from the base are transformed into that from a target distribution;
        we can analytically compute the likelihood of the target distribution using $g^{-1}$;
        by optimizing the invertible transformation(s), the desired distribution can be obtained.
    }
    \label{fig:nf}
\end{figure*}

The set of probability distributions $P$ for a random variable $x \in \mathcal{X}$ belonging to a continuous space is defined as a set of functions satisfying the following two conditions.
\begin{align}
P = \left \{ p \mid p(x) \geq 0, \ \int_\mathcal{X} p(x) dx = 1 \right \}
\label{eq:prob_set}
\end{align}
Learning $p^*$ that best represents a given problem is the basis of most machine learning, but to this end, it is necessary to restrict the model using a parameter $\theta$, which is optimized instead of $p$ directly.
With $\theta$, the subset $P_\theta \subset P$ is forcibly extracted, hence, if $p^* \notin P_\theta$, we no longer obtain the global optimal solution.
While normal distribution is widely used as the simplest model, its $\theta$ contains only the location and scale, and is with a small subset.

As a modeling method with higher expressiveness, normalizing flow (NF)~\cite{papamakarios2021normalizing} has attracted much attention in recent years.
Specifically, in NF, a simple base probability distribution $p_b(x_b)$ and a complex target probability distribution $p_t(x_t)$ are connected by an appropriate invertible transformation $g: x_b \to x_t$ ($g^{-1}: x_t \to x_b$) (see Fig.~\ref{fig:nf}).
In this case, $p_t(x_t)$ can be computed as follows:
\begin{align}
    p_t(x_t) = p_b(x_b) \left | \cfrac{\partial g(x_b)}{\partial x_b} \right |^{-1}
    = p_b(g^{-1}(x_t)) \left | \cfrac{\partial g^{-1}(x_t)}{\partial x_t} \right |
    \label{eq:flow_def}
\end{align}
That is, given $x_b$ or $x_t$, both $p_b$ and $p_t$ can be calculated, and maximization of the (log) likelihood is feasible.
Even if the parameter $\theta_b$ that determines $p_b$ is limited like normal distribution, the addition of $\theta_g$ related to the invertible transformation $g$ yields the improvement of expressiveness.
It is possible to extend the above formulation to conditional probability by making $\theta$ a function over condition $s$, $\theta(s)$.
In addition, $g$ can be naturally expressed as a composite of multiple invertible transformations.

%%%%%%%%%%%%%%%%%%%%%%
\subsection{Reinforcement learning}

Reinforcement learning (RL)~\cite{sutton2018reinforcement} aims to solve an optimal control problem that maximizes the sum of future rewards (named return) under an unknown environment and a trainable agent.
In general, Markov decision process is assumed, where one step can be represented by the current state $s$, action $a$, next state $s^\prime$, and reward $r$.
Specifically, the agent samples $a$ from a trainable policy over $s$, $\pi(a \mid s)$.
$a$ acts on the environment, then $s^\prime$ is obtained stochastically from the state transition probability $p_e(s^\prime \mid s, a)$, which is generally assumed to be invariant during (and even after) training.
At the same time, $r$ is obtained from the reward function $r(s, a, s^\prime)$, which suggests the goodness of the state and action.
The return from time step $t$ is given by $R_t = \sum_{k=0}^\infty \gamma^k r_{t+k}$ ($\gamma \in [0, 1)$ is the discount factor), and $\pi$ is optimized by repeating the above cycle to infer and maximize the expected value of $R_t$ (as the value function $V(s) = \mathbb{E}[R_t \mid s]$ and/or $Q(s,a) = \mathbb{E}[R_t \mid s, a]$).

In the above problem setting, it is not difficult to imagine that the design of $\pi$ has a significant impact on the solution obtained.
Here, we consider the problem with the explicit definition of $\pi$ as probability distribution, instead of Q-learning~\cite{mnih2015human}, which indirectly designs $\pi$ from $Q(s,a)$, or model-based RL~\cite{janner2019trust,aotani2021meta}, which learns $p_e$ and obtains the optimal $a$ through planning.
In this case, the loss function for optimizing $\pi$ can be given according to, such as advantage actor-critic (A2C) algorithm~\cite{mnih2016asynchronous} and soft actor critic (SAC)~\cite{haarnoja2018soft}.
Here, we introduce the loss function in A2C as follows:
\begin{align}
    \mathcal{L}_\pi &= - \mathbb{E}_{s \sim p_e, a \sim \pi}[A(s, a) \ln \pi(a \mid s)]
    \label{eq:opt_policy}
\end{align}
where $A(s,a) = Q(s,a) - V(s) = \mathbb{E}[r + \gamma V(s^\prime)] - V(s)$ denotes the advantage function, which approximates the temporal difference (TD) error.
Since the minimization of $\mathcal{L}_\pi$ can be interpreted as weighted log-likelihood maximization, NF introduced in the previous section can be applied to $\pi$.

%%%%%%%%%%%%%%%%%%%%%%%%%%%%%%%%%%%%%%%%%%%%%%%%%%%%%%%%%%%%%%%%%%%%%%%%%%%%%%%%
\section{Proposal}

%%%%%%%%%%%%%%%%%%%%%%
\subsection{Unstable control by stochastic behavior}
\label{subsec:unstable_control}

As mentioned in the introduction, one of the problems with NF is that its statistics cannot be obtained analytically.
Among them, the inability to obtain the mean can be a practical obstacle in RL.
When the trained policy is deployed to operate the robot as inference mode, action (a.k.a. control input) should be determined not as stochastic one by sampling from the policy, but as deterministic one based on the mean of the policy.
While stochastic behavior is useful for exploration during learning, it can also introduce risks such as task failure.

Indeed, we tested 100 trials of the RL benchmark task, Ant task implemented by Pybullet~\cite{coumans2016pybullet}, deploying the trained policy.
The worst-case score (the sum of the rewards) was halved to 980 when the sample from the policy was used, compared to 1928 when the mean of the policy was used.
As for the agent motion, the mean of the policy was able to continue walking, while the sample from the policy was unable to continue walking and stopped midway (see the attached video).
This is thought to be due to the stochastic behavior of the agent, causing it to visit unlearned states and not knowing the optimal action.

As in this one example, stochastic behavior is a source of concern in the operational phase.
To mitigate this concern and deploy a reliable controller, we design the policy model with a novel NF that has the analytic mean.
In addition, since this kind of design limits the expressiveness, we also discuss the ways to recover it.

%%%%%%%%%%%%%%%%%%%%%%
\subsection{Constraints for analytic mean}

%Figure
\begin{figure}[tb]
    \begin{subfigure}[b]{0.48\linewidth}
        \centering
        \includegraphics[keepaspectratio=true,width=\linewidth]{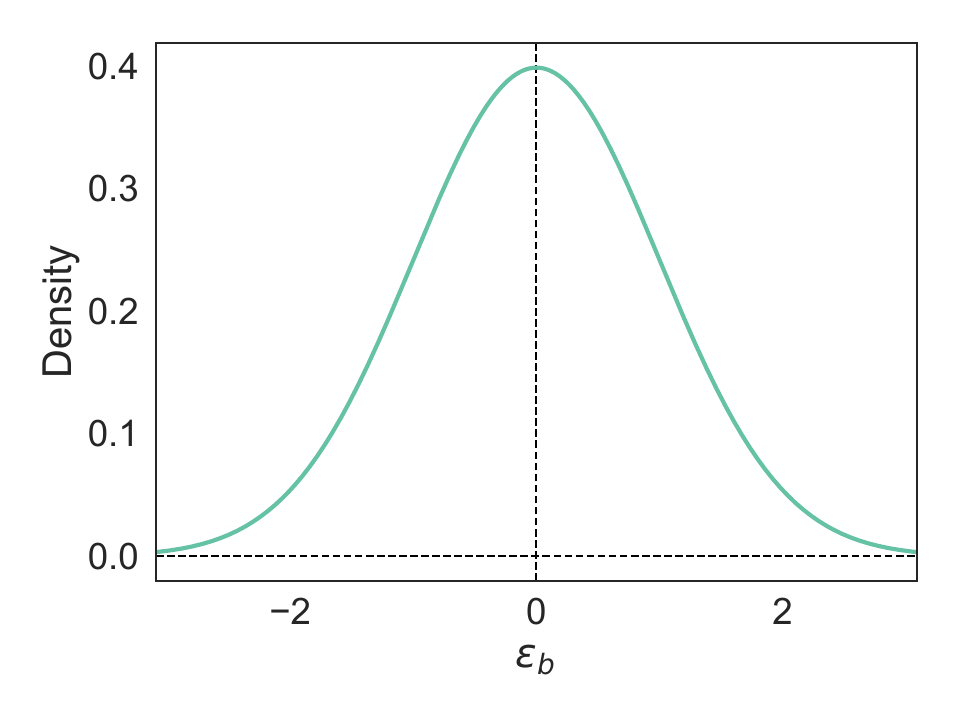}
        \subcaption{Symmetric distribution}
        \label{fig:const_sym}
    \end{subfigure}
    \begin{subfigure}[b]{0.48\linewidth}
        \centering
        \includegraphics[keepaspectratio=true,width=\linewidth]{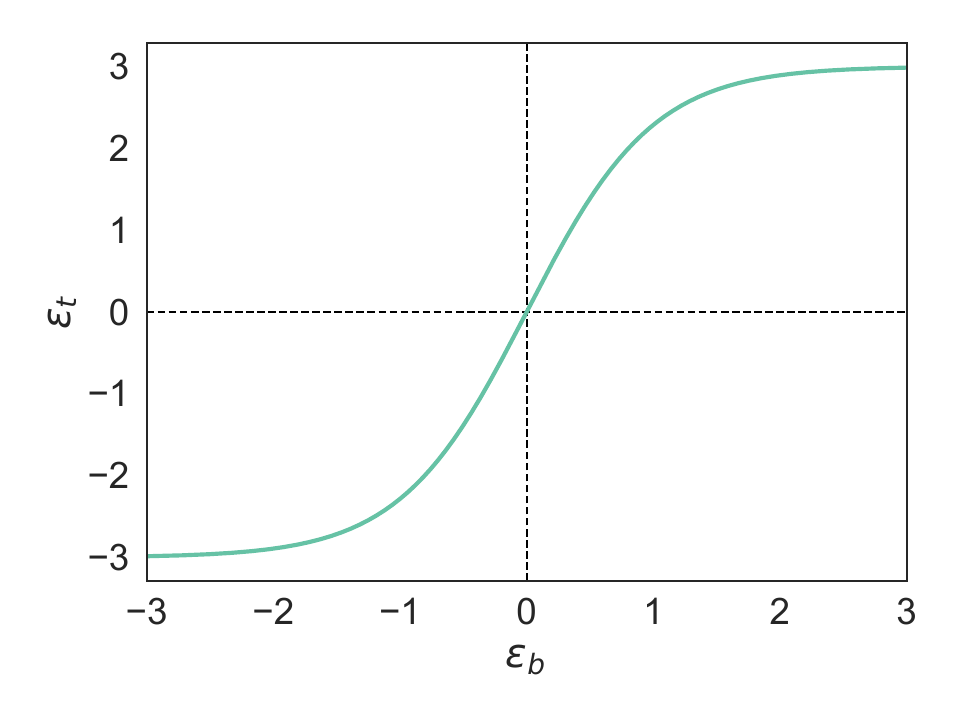}
        \subcaption{Odd transformation}
        \label{fig:const_odd}
    \end{subfigure}
    \caption{Constraints to make the mean of NF analytically computable:
        (a) the base distribution is limited to the symmetric one;
        (b) the invertible transformation is limited to the odd function.
    }
    \label{fig:const}
\end{figure}

First of all, we consider the mean of NF, $\mu_t$, as below.
\begin{align}
    \mu_t &= \int p_t(x_t) x_t dx_t
    \nonumber \\
    &= \int p_b(x_b) g(x_b) \left | \cfrac{\partial x_b}{\partial x_t} \right | dx_t
    \nonumber \\
    &= \int p_b(x_b) g(x_b) dx_b
    \label{eq:flow_mean1}
\end{align}
Even if $p_b$ is defined by a simple distribution model like normal distribution, the above integral involving the invertible transformation $g$ is generally not obtained in closed form.
Therefore, if $\pi$ is designed with NF, the mean of $\pi$ cannot be obtained analytically.

To resolve this problem, two design constraints are imposed (see Fig.~\ref{fig:const}).
One is the constraint for the base $p_b$ such that it is symmetric around the mean $\mu_b$, and $x_b$ can be sampled by the following equation together with its scale $\sigma_b$.
\begin{align}
    x_b = \mu_b + \sigma_b \epsilon_b
    \label{eq:norm_def}
\end{align}
where $\epsilon_b$ denotes the standardized random variable sampled from $p_{\epsilon_b}$, which is the distribution symmetric to the origin.
With this constraint, eq.~\eqref{eq:flow_mean1} can be further modified.
\begin{align}
    \int p_b(x_b) g(x_b) dx_b
    &= \int p_{\epsilon_b}(\epsilon_b) g(\mu_b + \sigma_b \epsilon_b) \left | \cfrac{\partial \epsilon_b}{\partial x_b} \right | dx_b
    \nonumber \\
    &= \int p_{\epsilon_b}(\epsilon_b) g(\mu_b + \sigma_b \epsilon_b) d\epsilon_b
    \label{eq:flow_mean2}
\end{align}

Another constraint is for $g$.
Specifically, $g$ is limited as follows:
\begin{align}
    g(\mu_b + \sigma_b \epsilon_b) &= \mu_t + \sigma_t \epsilon_t
    \label{eq:denorm_def} \\
    \epsilon_t &= g_\mathrm{odd}(\epsilon_b)
    \nonumber
\end{align}
where $g_\mathrm{odd}$ denotes the odd function for the invertible transformation of $\epsilon_b$.
$\mu_t$ and $\sigma_t$ are transformed from $\mu_b$ and $\sigma_b$ using such as affine transformation, which can be regarded as one of the invertible transformation in NF, although this paper simplifies $\mu_t = \mu_b$ and $\sigma_t = \sigma_b$ without the affine transformation.
That is, $g$ is limited to the odd function.
With this constraint, eq.~\eqref{eq:flow_mean2} can be eventually solved.
\begin{align}
    &\int p_{\epsilon_b}(\epsilon_b) g(\mu_b + \sigma_b \epsilon_b) d\epsilon_b
    \nonumber \\
    &= \mu_t \int p_{\epsilon_b}(\epsilon_b) d\epsilon_b
    + \sigma_t \int p_{\epsilon_b}(\epsilon_b) g_\mathrm{odd}(\epsilon_b) d\epsilon_b
    \nonumber \\
    &= \mu_t
    \label{eq:flow_mean3}
\end{align}
In the second line, the first term can be computed as one according to eq.~\eqref{eq:prob_set}, and the second term can be eliminated due to the integral of the odd function with symmetric $p_{\epsilon_b}$ (i.e. the even function) and the odd $g_\mathrm{odd}$.

In summary, the analytical solution for $\mu_t$ is obtained by restricting the base $p_b$ to a symmetric distribution and the invertible transformation $g$ to an odd function acting on the standardized random variable $\epsilon_b$ (and an affine transformation from $\mu_b, \sigma_b$ to $\mu_t, \sigma_t$, respectively).
As a remark, the mean computation requires almost no complex computation in NF, the computational cost is theoretically almost the same as that of the base mean.
With this restricted NF (RNF), we can design $\pi$ with the analytic mean, high expressiveness, and low computational cost.

%%%%%%%%%%%%%%%%%%%%%%
\subsection{Example}

The above constraints, in particular $g_\mathrm{odd}$, are general definitions, and specific implementations need to be given for each of the conventional NF methods.
In this section, we introduce two representative examples:
LRS flow~\cite{dolatabadi2020invertible} as a spline type NF (see below);
and RealNVP~\cite{dinh2016density} as an affine type NF (in Appendix~\ref{app:realnvp}).
A concrete $g_\mathrm{odd}$ for each is then introduced.
Note that, since the dimensionality of the action space in RL can be one, LRS flow, which assumes transformation for each axis unit, is suitable for this study instead of methods that assume multiple dimensions such as RealNVP.

%Figure
\begin{figure}[tb]
    \begin{subfigure}[b]{0.48\linewidth}
        \centering
        \includegraphics[keepaspectratio=true,width=\linewidth]{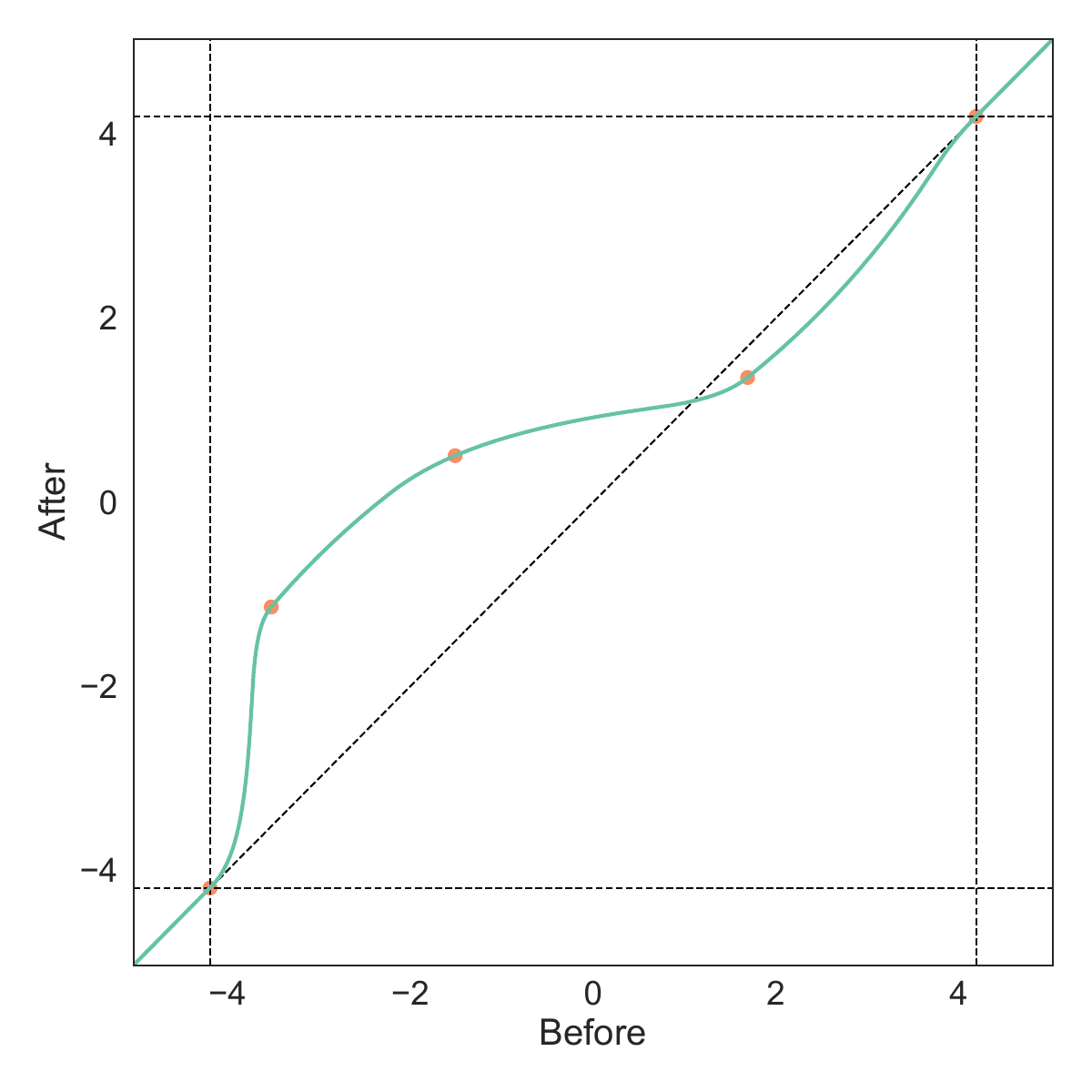}
        \subcaption{Original version}
        \label{fig:lrs_basic}
    \end{subfigure}
    \begin{subfigure}[b]{0.48\linewidth}
        \centering
        \includegraphics[keepaspectratio=true,width=\linewidth]{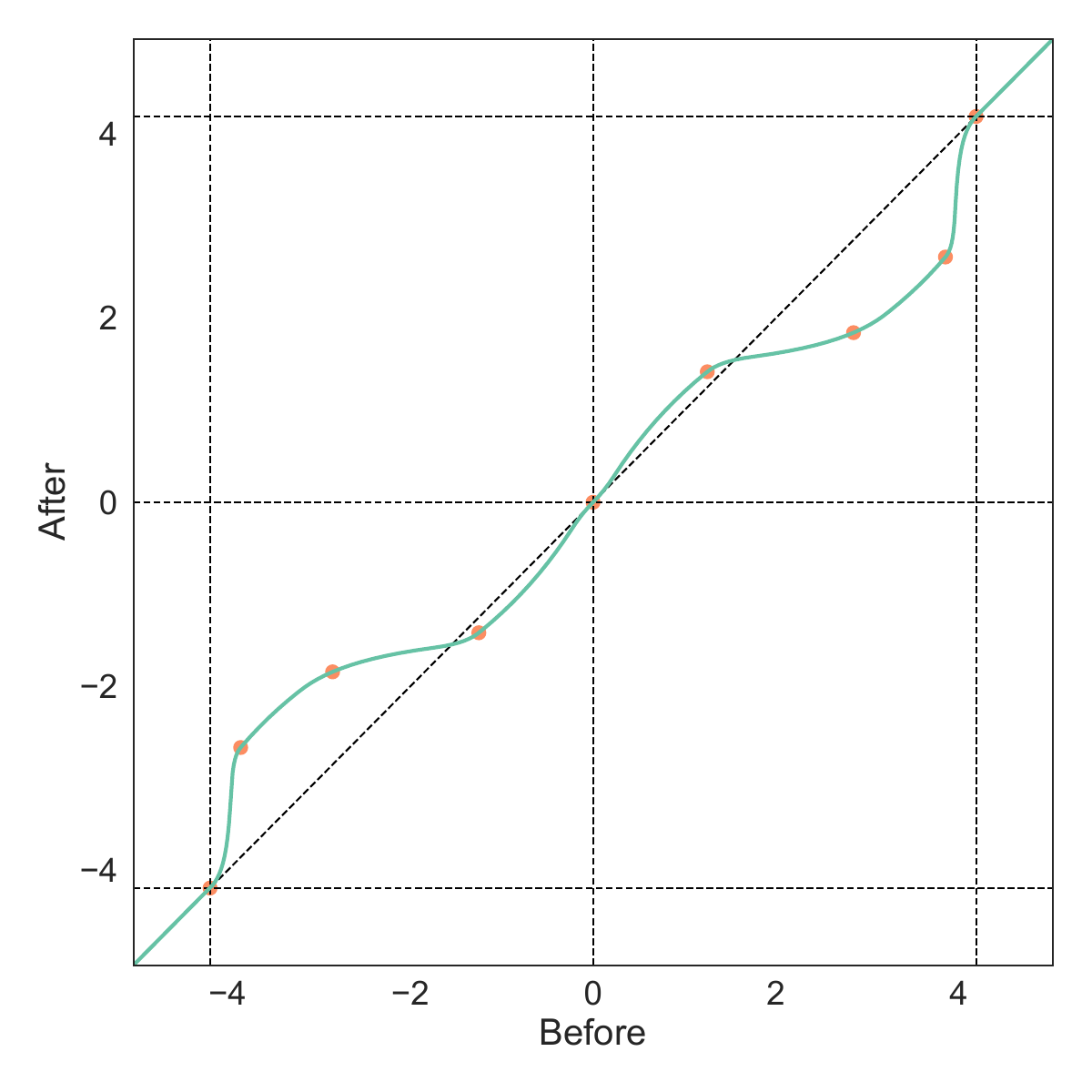}
        \subcaption{Restricted version}
        \label{fig:lrs_restricted}
    \end{subfigure}
    \caption{LRS flow:
        (a) the random variable from the base is smoothly transformed to the target one in the specified interval with LRS;
        (b) the inverted first half of the interval is placed and combined with the second half so that LRS becomes an odd function.
    }
    \label{fig:lrs}
\end{figure}

The basic LRS flow~\cite{dolatabadi2020invertible} is first introduced.
It applies a monotonically increasing spline transformation to each axis passing through $K+1$ ($K \in \mathbb{N}$) knots in the transformation interval $[l, u]$ (see Fig.~\ref{fig:lrs_basic}).
The parameter $\theta_g$ contains three types at each knot $k = 0,1,2,\ldots,K$:
the normalized position of $k$-th knot, $(q_b^k \in [0, 1], q_t^k \in [0, 1])$ ($q_{b,t}^{k-1} < q_{b,t}^k$);
the interpolation placed between $k-1$-th and $k$-th knots (except $k=0$), $\phi_m^k \in (0, 1)$;
and the slope at $k$-th knot, $d^k > 0$.
As necessary conditions, the first and last knots must be placed at the beginning and end of the interval (i.e. $q_{b,t}^0 = 0$ and $q_{b,t}^K = 1$).
In addition, the slope at the beginning and end of the interval must be $1$ to guarantee continuity with the outside of the interval (i.e. $d^{0,K} = 1$).

The transformation process consists of five steps, $x_b \to q_b \to \phi \to \psi \to q_t \to x_t$, but except for $\phi \to \psi$, they can be regarded as the encoding/decoding processes.
\begin{align}
    q_b(x_b) &= \cfrac{x_b - l}{u - l}
    \label{eq:xb2qb} \\
    \phi(q_b) &= \cfrac{q_b - q_b^k}{q_b^{k+1} - q_b^k}
    \label{eq:qb2phi} \\
    q_t(\psi) &= (q_t^{k+1} - q_t^k) \psi + q_t^k
    \label{eq:psi2qt} \\
    x_t(q_t) &= (u - l) q_t + l
    \label{eq:qt2xt}
\end{align}
where $k$ is chosen to satisfy $q_b^k \leq q_b < q_b^{k+1}$.
Note that, from the original paper~\cite{dolatabadi2020invertible}, $x_b \to q_b$ and $q_t \to x_t$ are additionally applied for fully use of the transformation interval.

The remaining $\phi \to \psi$ is the main transformation by LRS as below.
\begin{align}
    \psi(\phi) =
    \begin{cases}
        \cfrac{\psi_m^k w_m^k \phi}{w_m^k \phi + w_s^k (\phi_m^k - \phi)}
        & 0 \leq \phi < \phi_m^k
        \\
        \cfrac{\psi_m^k w_m^k (1 - \phi) + w_e^k (\phi - \phi_m^k)}{w_m^k (1 - \phi) + w_e^k(\phi - \phi_m^k)}
        & \phi_m^k \leq \phi < 1
    \end{cases}
    \label{eq:phi2psi}
\end{align}
where
\begin{align}
    w_s^k &= 1 / \sqrt{d^k}
    \label{eq:wsk} \\
    w_e^k &= 1 / \sqrt{d^{k+1}}
    \label{eq:wek} \\
    w_m^k &= (\phi_m^k / w_s^k + (1 - \phi_m^k) / w_e^k) \cfrac{q_b^{k+1} - q_b^k}{q_t^{k+1} - q_t^k}
    \label{eq:wmk} \\
    \psi_m^k &= \cfrac{\phi_m^k}{w_s^k w_m^k} \cfrac{q_b^{k+1} - q_b^k}{q_t^{k+1} - q_t^k}
    \label{eq:psimk}
\end{align}
Note that the inverse transformation, $x_t \to x_b$, coincides with the replacement of $x_b$ and $x_t$, $\phi$ and $\psi$, $q_b^k$ and $q_t^k$, $\phi_m^k$ and $\psi_m^k$, $w_s^k$ and $w_m^k$, and $w_e^k$ and $w_m^k$, respectively.

The gradient for each axis can be analytically computed as follows:
\begin{align}
    \cfrac{\partial x_t}{\partial x_b} &= \cfrac{q_t^{k+1} - q_t^k}{q_b^{k+1} - q_b^k}
    \nonumber \\
    &\times
    \begin{cases}
        \cfrac{w_m^k w_s^k \psi_m^k \phi_m^k}{\{w_m^k \phi + w_s^k (\phi_m^k - \phi)\}^2}
        & 0 \leq \phi < \phi_m^k
        \\
        \cfrac{w_m^k w_e^k (1 - \psi_m^k) (1 - \phi_m^k)}{\{w_m^k (1 - \phi) + w_e^k(\phi - \phi_m^k)\}^2}
        & \phi_m^k \leq \phi < 1
    \end{cases}
    \label{eq:jacobian_lrs}
\end{align}
The determinant for the entire transformation is the product of the above equations, which are the gradients for each axis, because the gradients are diagonal.
Note that the above gradient should take into account the transformation by eqs.~\eqref{eq:norm_def} and~\eqref{eq:denorm_def}, but their gradients are canceled by each other when $\sigma_t = \sigma_b$.

For the above LRS flow, we consider a specific design of $\theta_g$ that satisfies the constraints.
First, as shown in eq.~\eqref{eq:norm_def}, since $\epsilon_b$ is a standardized random variable and is symmetric to the origin, the interval of LRS Flow should be defined to be symmetric $[-c, c]$.
With this symmetric interval, to get $g_\mathrm{odd}$, we have to set $\epsilon_t = 0$ when $\epsilon_b = 0$.
In this case, it is easy to design $g_\mathrm{odd}$ such that the inverted first half of the interval is placed and combined with the second half so that LRS becomes an odd function (see Fig.~\ref{fig:lrs_restricted}).

Specifically, $\theta_g$ can be constrained as follows:
\begin{align}
    \boldsymbol{q}_{b,t} \gets \cfrac{1}{2} \left [ \boldsymbol{q}_{b,t}[:-1]^\top, 1 + \bar{\boldsymbol{q}}_{b,t}^\top \right ]^\top
    ,
    \boldsymbol{\phi}_m \gets \left [ \boldsymbol{\phi}_m^\top, 1 - \bar{\boldsymbol{\phi}}_m^\top \right ]^\top
    ,
    \boldsymbol{d} \gets \left [ \boldsymbol{d}[:-1]^\top, \bar{\boldsymbol{d}}^\top \right ]^\top
    \label{eq:lrs_extend}
\end{align}
where $\bar{\boldsymbol{x}}$ indicates an operation that reverses the order of the elements in a vector, and $\boldsymbol{x}[:-1]$ indicates a subvector that excludes the final element.
Only with the above modification, the transformation process by eqs.~\eqref{eq:xb2qb}--\eqref{eq:phi2psi} yields $g_\mathrm{odd}$.

%%%%%%%%%%%%%%%%%%%%%%
\subsection{Numerical stability}

It should be noted that NF has been reported to be unstable~\cite{behrmann2021understanding}, so we introduce a hyperparameter $\tau \in (0, 1)$ that specifies the capability of transformation and stabilizes the numerical computation of NF.
Specifically, in order for the invertible transform $g$ of (R)NF to always produce stable outputs, it must be bi-Lipschitz continuous~\cite{behrmann2021understanding}.
This is only possible if both $\partial g(x_b) / \partial x_b$ and $\partial g^{-1}(x_t) / \partial x_t$ are bounded as $(1/\kappa, \kappa)$ ($\kappa > 0$, but a large amount is desirable for numerical stability).
A naive implementation that satisfies this would be to restrict the domain of definition of the parameter $\theta_g$ using $\tau \in (0, 1)$.

For example, in LRS flow (even with the above restriction), $q_{b,t}$, $\phi_m$, and $d$ are bounded, respectively.
As for $q_{b,t}$, we consider $\Delta q_{b,t}$ as the deviation between knots, and restrict its aspect ratio $r = \Delta q_t / \Delta q_b$ within $(1-\tau, 1/(1 - \tau))$.
That is, we directly design $\Delta q_{b,t}$ that satisfies $r \in (1-\tau, 1/(1 - \tau))$, then obtain $q_{b,t}$ from the accumulation of $\Delta q_{b,t}$.
$\phi_m$, which represents the relay point between the knots, should be placed near the center since if it is too close to either knot, its gradient is prone to be too extreme.
Therefore, we design $\phi_m \in ((1 - \tau)/2, (1 + \tau)/2)$
As for $d$, the slope at the knots, it is simply given to be $d \in (1-\tau, 1/(1 - \tau))$.
These designs can be implemented using $y^{*}$ the variable mapped onto the real-valued space for each parameter.
\begin{align}
    d &= \exp\{\ln(1 - \tau) \mathrm{squaresign}(y^d)\}
    \\
    \phi_m &= \cfrac{1}{2} \{ 1 + \mathrm{squaresign}(y^{\phi_m})\}
    \\
    \Delta q_{b,t} &= \cfrac{\mathrm{squaremax}(y^{\Delta q_{b,t}}) + (1 - \tau) / \tau}{1 + K (1 - \tau) / \tau}
\end{align}
where $\mathrm{squaresign}$ and $\mathrm{squaremax}$ are introduced for fast and stable computation (see Appendix~\ref{app:nonlinear}).

In addition, two hyperparameters in LRS flow, $K$ and $c$, are also specified according to $\tau$.
Focusing on the fact that $K$ was given as a maximum of $32 = 2^5$ in the original paper~\cite{dolatabadi2020invertible}, we define $K = \lfloor 2^{5\tau} \rceil$.
Similarly, $c$ was mostly given to be $5$ at maximum, we can simply set $c = 5 \tau$.

%%%%%%%%%%%%%%%%%%%%%%
\subsection{Heuristic solutions for inexpressible distributions}

While the constraints in RNF makes it possible to calculate the analytic mean, it loses the rich expressiveness for representing arbitrary probability distributions.
In this section, we describe the types of inexpressible distributions and the solutions for representing them again.

%%%%%%%%%%%%%%%%%%%%%%
\subsubsection{Asymmetric distribution}

%Figure
\begin{figure}[tb]
    \centering
    \includegraphics[keepaspectratio=true,width=0.96\linewidth]{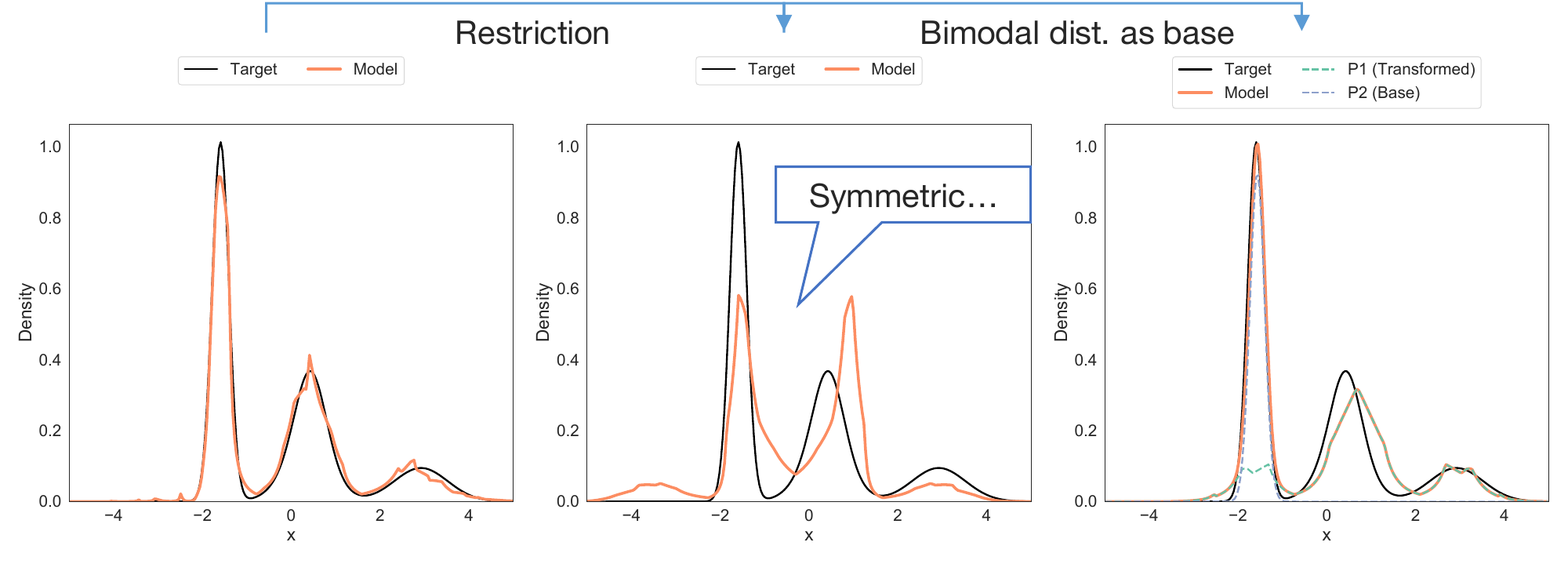}
    \caption{Reverting the expressiveness of asymmetric distributions:
        RNF transforms the base symmetric distribution to the complicated symmetric distribution;
        by adding another base distribution as bimodal distribution, it enables to represent the asymmetric distributions.
    }
    \label{fig:rnf_bimodal}
\end{figure}

Since RNF performs an invertible transformation of a symmetric base with an odd function, the resulting probability distribution must be symmetric, as shown in the left and center of Fig.~\ref{fig:rnf_bimodal}.
Although asymmetrization-specific transformations (e.g.~\cite{foulloy2021new}) have been proposed, to the best of our knowledge, none can derive the analytic mean in arbitrary case.

As an alternative solution, the base is given to be a bimodal mixture distribution with a symmetric distribution to be transformed and another distribution whose mean is known, $p_a$.
\begin{align}
    p(x) = \rho p_t(x) + (1 - \rho) p_a(x)
    \label{eq:bimodal}
\end{align}
where $\rho \in [0, 1]$ denotes the (trainable) mixture ratio.
This composition allows for symmetry breaking (see the right of Fig.~\ref{fig:rnf_bimodal}), and the composite mean is also obtained as a weighted mean of the analytical means, $\mu_t$ and $\mu_a$, respectively.
\begin{align}
    \mu &= \int_{\mathcal{X}} p(x) x dx
    \nonumber \\
    &= \rho \int_{\mathcal{X}} p_t(x) x dx + (1 - \rho) \int_{\mathcal{X}} p_a(x) x dx
    \nonumber \\
    &= \rho \mu_t + (1 - \rho) \mu_a
\end{align}
Note that limiting the number of mixtures to two minimizes the increase in computational cost.

%%%%%%%%%%%%%%%%%%%%%%
\subsubsection{Heavy-tailed distribution}

%Figure
\begin{figure}[tb]
    \centering
    \includegraphics[keepaspectratio=true,width=0.96\linewidth]{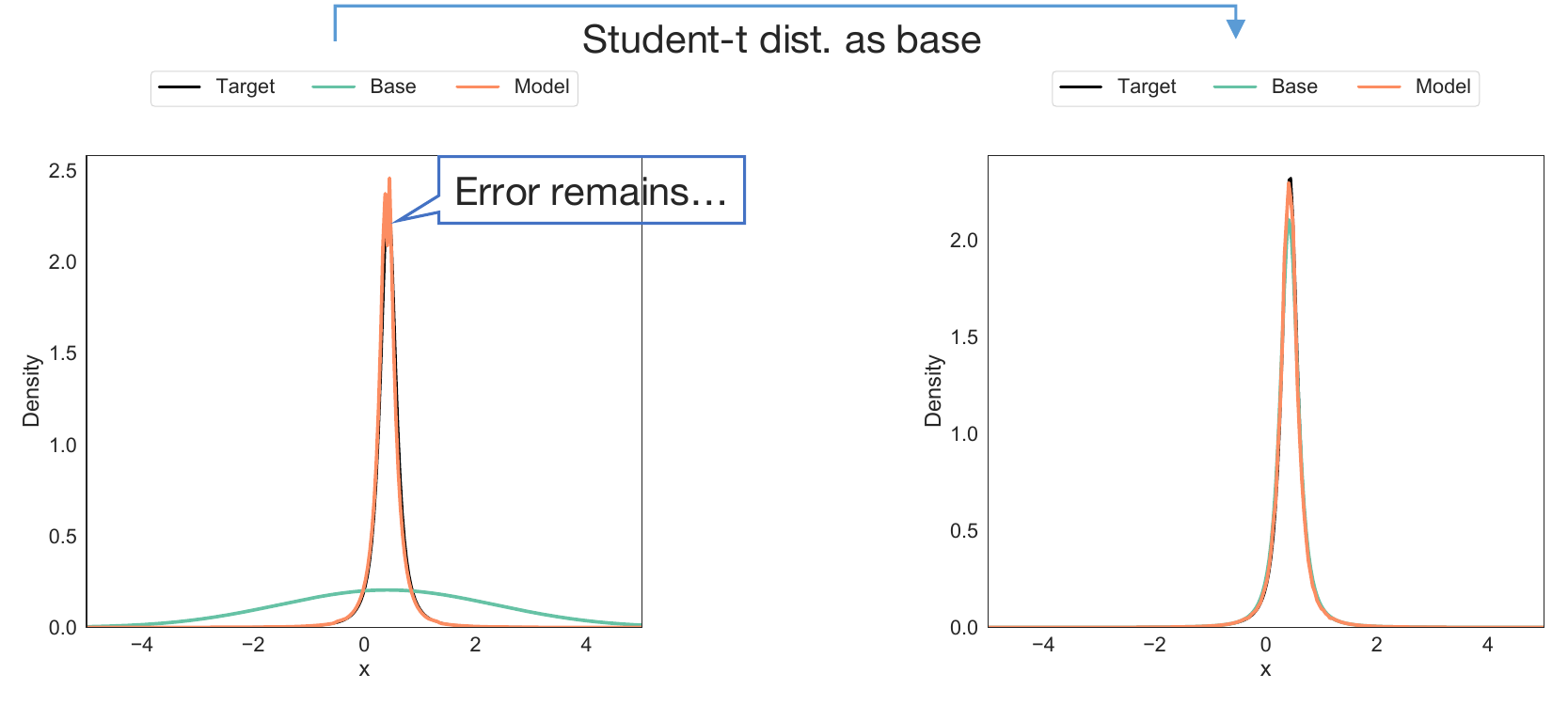}
    \caption{Reverting the expressiveness of heavy-tailed distributions:
        spline-type transformations like LRS and restrictions to stabilize numerical calculation cause the lack of the expressiveness for heavy tail, and errors remain as a side effect;
        by introducing an (adjustable) heavy-tailed distribution as the base, a heavy-tailed distribution can be naturally expressed without an unreasonable transformation.
    }
    \label{fig:rnf_t}
\end{figure}

Spline-type NFs such as LRS flow define a transformation interval, so distribution shapes outside that interval cannot be transformed.
When the transformation interval changes according to the base scale, as in the proposed implementation, it is possible to transform the tail shape to some extent.
However, if the base scale is extremely large to cover the heavy tail, such an intractable transformation is prone to leave errors and/or cause numerical instability (see the left of Fig.~\ref{fig:rnf_t}).
While affine-transformed NFs such as RealNVP have no specified transformation interval, the restriction of expressiveness for numerical stability still places a limit on the transformation to the tail.

A simple solution to this problem would be to change the base distribution as above.
Specifically, the base can be designed as a distribution with adjustable tail shape, such as student-t distribution~\cite{kobayashi2019student} (or its further extension q-Gaussian~\cite{umarov2008q}).
This simple extension eliminates the errors with a reasonable transformation, as shown in the right of Fig.~\ref{fig:rnf_t}, where the base guarantees the tail expression.

In this paper, we implement and test RNF with bimodal student-t distribution as the base, so-called Bit-RNF, according to the above two ideas.

%%%%%%%%%%%%%%%%%%%%%%
\subsubsection{Non-diagonal distribution}

Finally, the expressiveness for the inter-axial dependencies is limited.
Specifically, only the rotational and shear transformations that occur during the affine transformation to $\mu_b,\sigma_b \to \mu_t,\sigma_t$ can be available.
Other curved (e.g. banana-shaped) distribution cannot be obtained.
Since $\mu_b,\sigma_b$ has information on each axis but has nothing to do with the integral for computing the analytic mean, it can be given to the RNF condition, but it cannot transform the distribution shape as described above.

One solution is to make the base space highly independent.
Note that this paper does not directly address this issue since the RL action space can already be considered highly independent.

%%%%%%%%%%%%%%%%%%%%%%%%%%%%%%%%%%%%%%%%%%%%%%%%%%%%%%%%%%%%%%%%%%%%%%%%%%%%%%%%
\section{Experiment}

%%%%%%%%%%%%%%%%%%%%%%
\subsection{Setup}

%Table
\begin{table}[tb]
    \caption{Hyperparameters for the used RL algorithms}
    \label{tab:param}
    \centering
    \begin{tabular}{ccc}
        \hline\hline
        Symbol & Meaning & Value
        \\
        \hline
        $\gamma$ & Discount factor & $0.99$
        \\
        $(\alpha, \beta, \epsilon, \underline{\tilde{\nu}})$ & For AdaTerm~\cite{ilboudo2022adaterm} & $(10^{-4}, 0.9, 10^{-5}, 1)$
        \\
        $(\kappa, \beta, \lambda, \underline{\Delta})$ & For PPO-RPE~\cite{kobayashi2021proximal,kobayashi2022proximal} & $(0.5, 0.5, 0.999, 0.1)$
        \\
        $(N_c, N_b, \alpha, \beta)$ & For PER~\cite{schaul2015prioritized} & $(10^4, 32, 1.0, 0.5)$
        \\
        $(\tau, \nu)$ & For t-soft update~\cite{kobayashi2021t} & $(0.1, 1.0)$
        \\
        $(\beta_\mathrm{ent}, \beta_\mathrm{td})$ & For reward bonus~\cite{haarnoja2018soft,parisi2019td} & $(0.015, 0.005)$
        \\
        \hline\hline
    \end{tabular}
\end{table}

We describe the specific setup for comparative validation.
Hyperparameters are summarized in Table~\ref{tab:param}.

First, the policy and value functions, $\pi$ and $V$, required in RL are approximated by DNNs with PyTorch~\cite{paszke2017automatic}.
Their network architecture takes the state $s$ as input, 5 fully-connected networks (FCNs) (with 100 neurons for each) as hidden layers, and linearly transforms the features obtained there to the output.
Layer normalization~\cite{ba2016layer} and squish function (see Appendix~\ref{app:nonlinear}) are combined as activation functions for FCNs.
Five models for $\pi$ are compared (see details of them in Appendix~\ref{app:distribution}):
normal distribution as \textit{Normal};
student-t distribution as \textit{Student};
GMM with 10 components as \textit{GMM-10} (almost same computational cost as Bit-RNF below);
GMM with 16 components as \textit{GMM-16} (same as the number of knots, $K$, in Bit-RNF below);
and the proposed \textit{Bit-RNF} with $\tau = 0.8$.
Note that we consider RNF conditional on the features obtained by FCNs, and transform the features into $\theta_g$ in a small network with two FCNs (with 32 neurons for each) and the squaresign function (see Appendix~\ref{app:nonlinear}) as the activation function.
This design is based on the literature that even this size has sufficient expressiveness.
To improve learning and inference accuracy of $V$, an ensemble of multiple (specifically, five) $V$ outputs is employed, as in the literature~\cite{osband2016deep}.
The above network is updated by AdaTerm~\cite{ilboudo2022adaterm}, which is a type of stochastic gradient descent method that is robust to noise in the training signal.

PPO-RPE~\cite{kobayashi2021proximal,kobayashi2022proximal} is employed as the base of RL algorithm.
$\pi$ and $V$ are stabilized by introducing target networks (updated by t-soft update~\cite{kobayashi2021t}), and the empirical data are efficiently reused by prioritized experience replay (PER)~\cite{schaul2015prioritized}.
In addition, as a bonus term to the reward, a term to maximize entropy as in SAC~\cite{haarnoja2018soft}, $-\ln \pi$, and a term to suppress TD error~\cite{parisi2019td}, $|r + \gamma V(s^\prime) - V(s)|$, are added with the respective gains.

%%%%%%%%%%%%%%%%%%%%%%
\subsection{Benchmark}
\label{subsec:benchmark}

%Figure
\begin{figure}[tb]
    \begin{subfigure}[b]{0.48\linewidth}
        \centering
        \includegraphics[keepaspectratio=true,width=\linewidth]{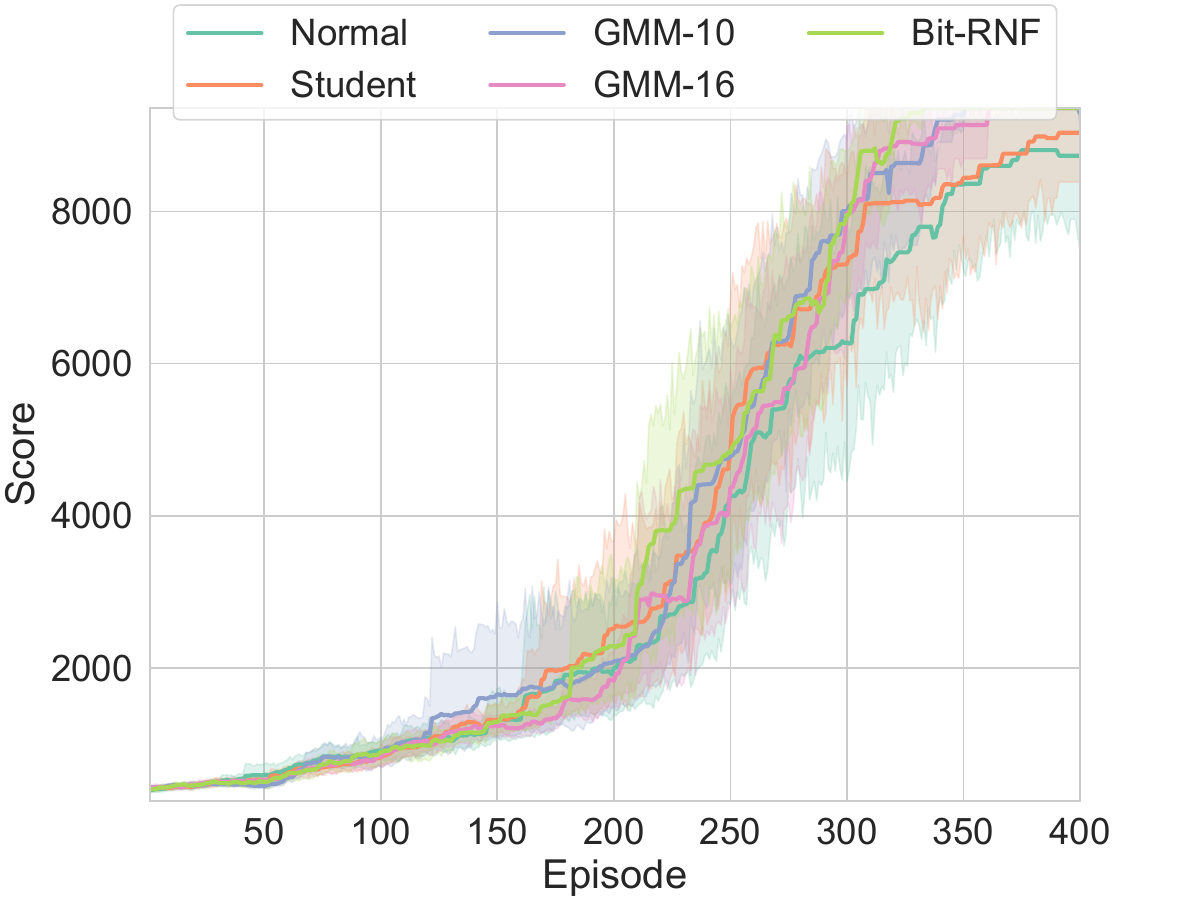}
        \subcaption{DoublePendulum}
        \label{fig:sim_score_DoublePendulum}
    \end{subfigure}
    \begin{subfigure}[b]{0.48\linewidth}
        \centering
        \includegraphics[keepaspectratio=true,width=\linewidth]{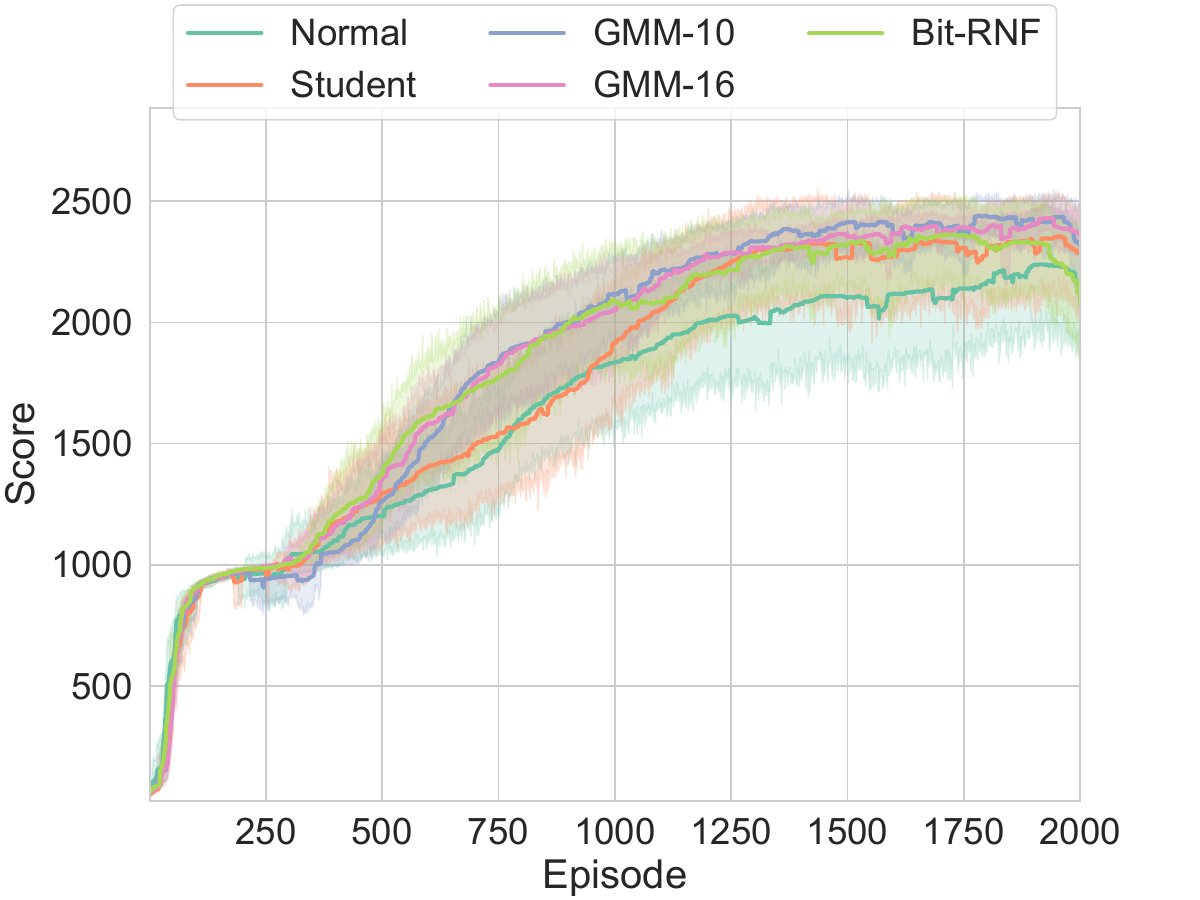}
        \subcaption{Hopper}
        \label{fig:sim_score_Hopper}
    \end{subfigure}
    \begin{subfigure}[b]{0.48\linewidth}
        \centering
        \includegraphics[keepaspectratio=true,width=\linewidth]{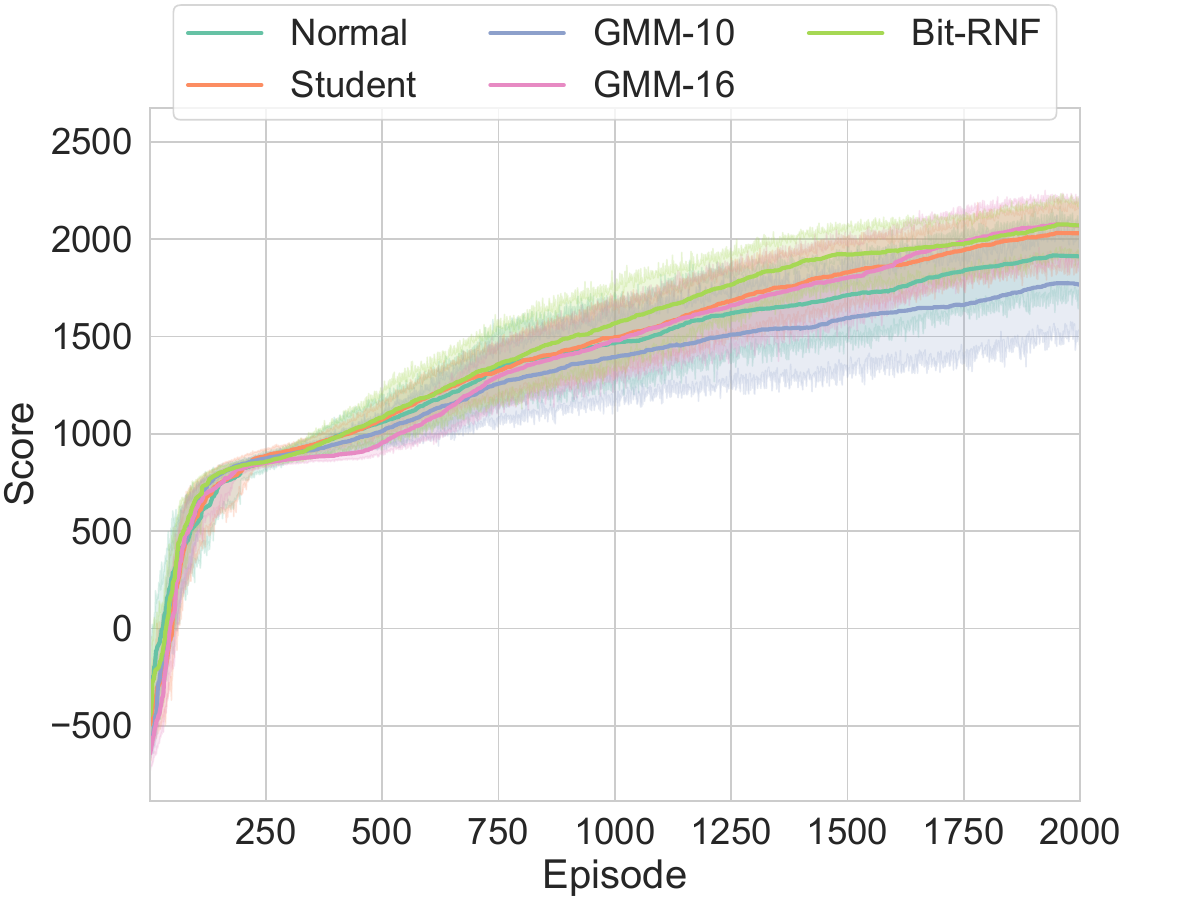}
        \subcaption{HalfCheetah}
        \label{fig:sim_score_HalfCheetah}
    \end{subfigure}
    \begin{subfigure}[b]{0.48\linewidth}
        \centering
        \includegraphics[keepaspectratio=true,width=\linewidth]{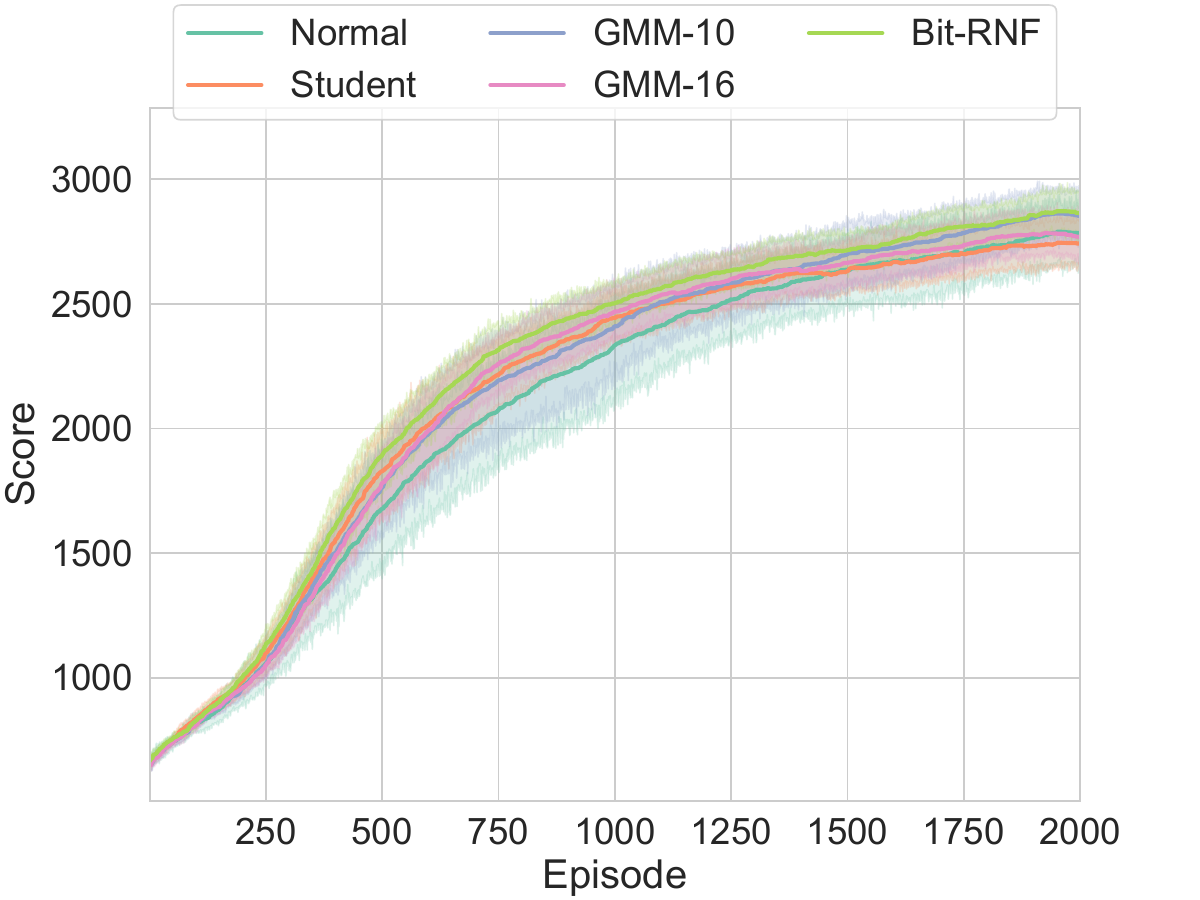}
        \subcaption{Ant}
        \label{fig:sim_score_Ant}
    \end{subfigure}
    \caption{Learning curves for the respective benchmark tasks:
        the representative model, normal distribution, lagged behind other models in learning;
        despite the addition of only one parameter $\nu$, student-t distribution generally outperformed normal distribution;
        GMM often failed to learning HalfCheetah, although gained high-level performance in the other tasks;
        the scores of Bit-RNF increased early in all tasks, indicating high sample efficiency.
    }
    \label{fig:sim_score}
\end{figure}

%Table
\begin{table}[tb]
    \caption{The sum of rewards after training:
        the mean and standard deviation for the 100 tests are listed;
        for each test, the analytic mean of the policy was used as action;
        successful cases are noted in bold.
    }
    \label{tab:sim_score}
    \centering
    \begin{tabular}{l cccc}
        \hline\hline
        Method & DoublePendulum & Hopper & HalfCheetah & Ant
        \\
        \texttt{reward\_threshold}
        & 9100
        & 2500
        & 3000
        & 2500
        \\
        \textit{Success}
        & 8000
        & 2000
        & 2000
        & 2000
        \\
        \hline
        Normal
        & 6254 $\pm$ 3440
        & 1252 $\pm$ 811
        & 1970 $\pm$ 470
        & \textbf{2820} $\pm$ 328
        \\
        Student
        & \textbf{8245} $\pm$ 2280
        & 1906 $\pm$ 922
        & \textbf{2082} $\pm$ 389
        & \textbf{2751} $\pm$ 226
        \\
        GMM-10
        & \textbf{8073} $\pm$ 2758
        & \textbf{2227} $\pm$ 515
        & 1666 $\pm$ 938
        & \textbf{2864} $\pm$ 265
        \\
        GMM-16
        & \textbf{8087} $\pm$ 2532
        & \textbf{2370} $\pm$ 124
        & \textbf{2147} $\pm$ 408
        & \textbf{2829} $\pm$ 186
        \\
        Bit-RNF
        & \textbf{8586} $\pm$ 2037
        & \textbf{2199} $\pm$ 583
        & \textbf{2119} $\pm$ 321
        & \textbf{2934} $\pm$ 236
        \\
        \hline\hline
    \end{tabular}
\end{table}

%Figure
\begin{figure}[tb]
    \begin{subfigure}[b]{0.32\linewidth}
        \centering
        \includegraphics[keepaspectratio=true,width=\linewidth]{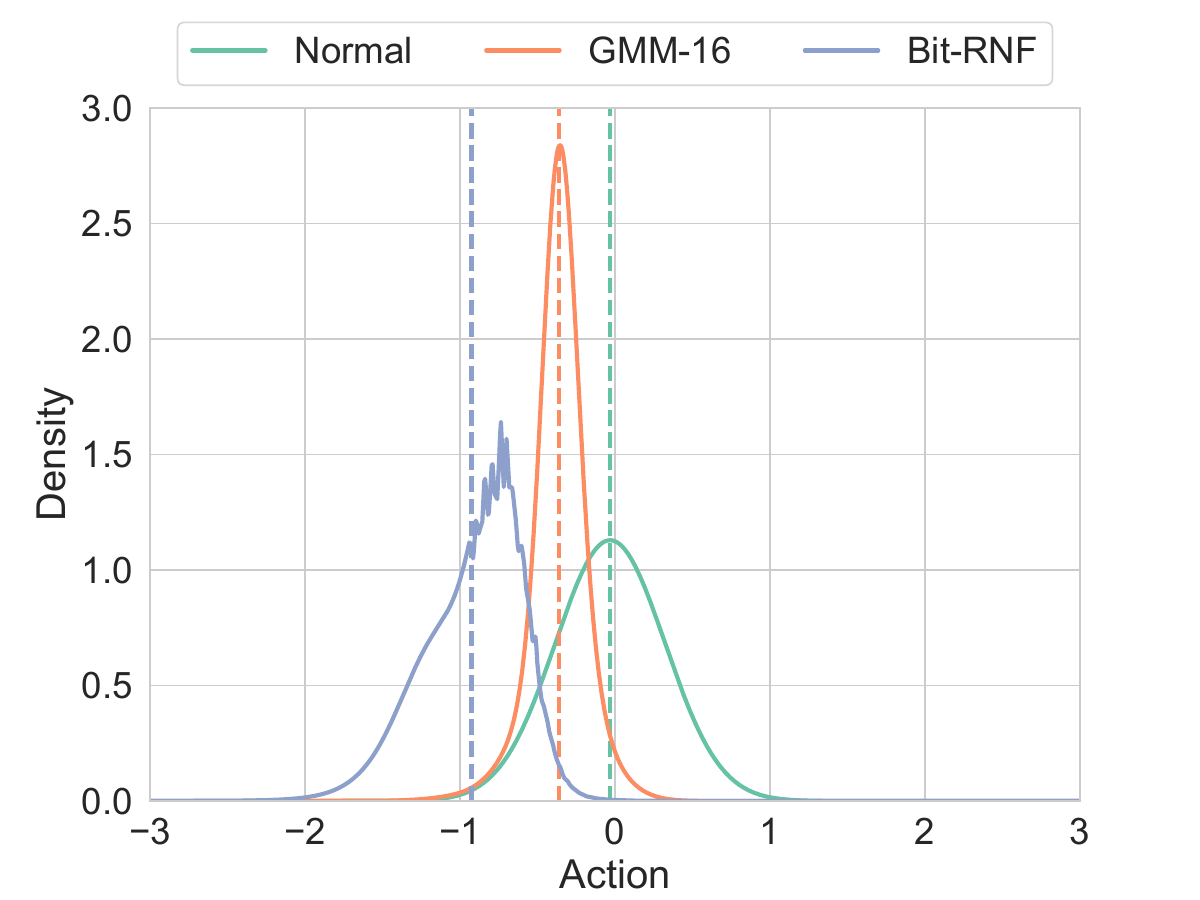}
        \subcaption{Asymmetric shape}
        \label{fig:example_policy_asymmetric}
    \end{subfigure}
    \begin{subfigure}[b]{0.32\linewidth}
        \centering
        \includegraphics[keepaspectratio=true,width=\linewidth]{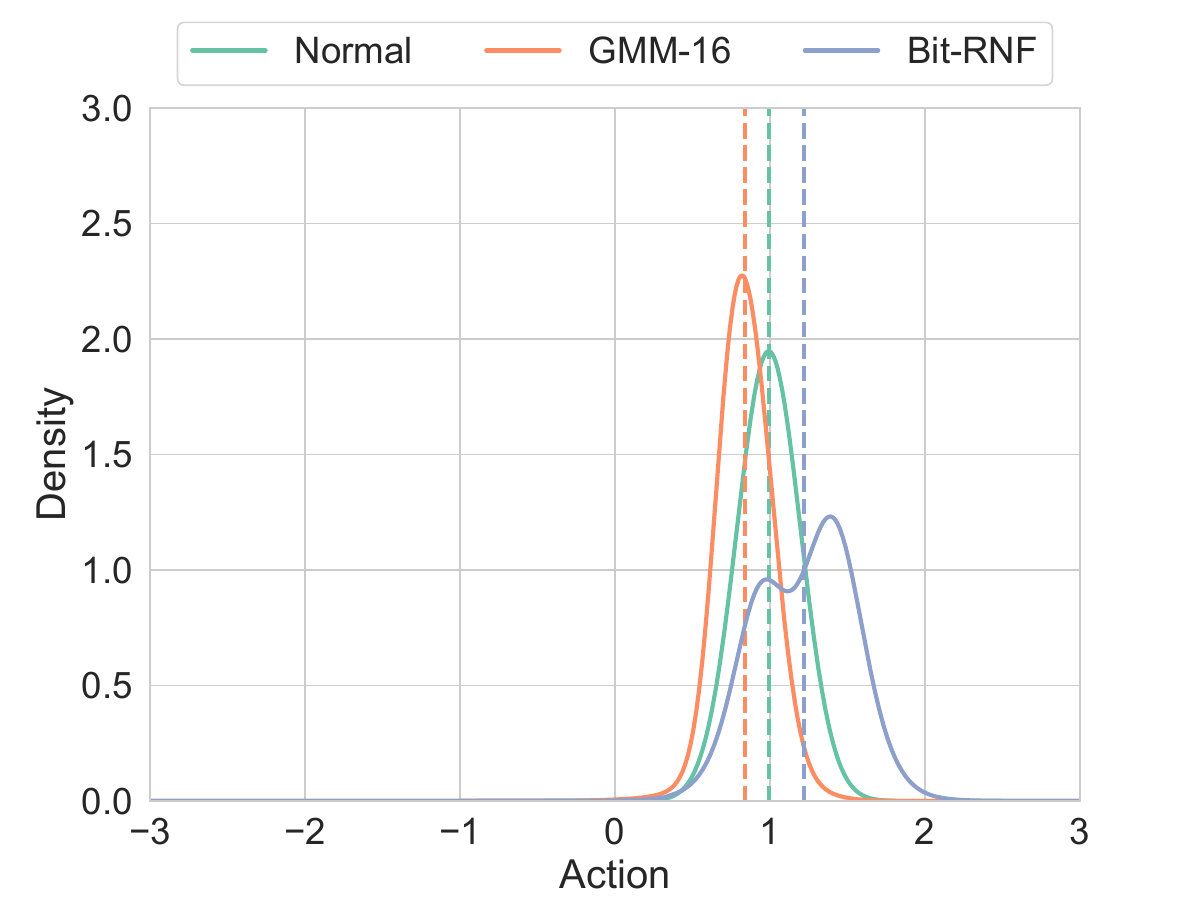}
        \subcaption{Multimodal shape}
        \label{fig:example_policy_multimodal}
    \end{subfigure}
    \begin{subfigure}[b]{0.32\linewidth}
        \centering
        \includegraphics[keepaspectratio=true,width=\linewidth]{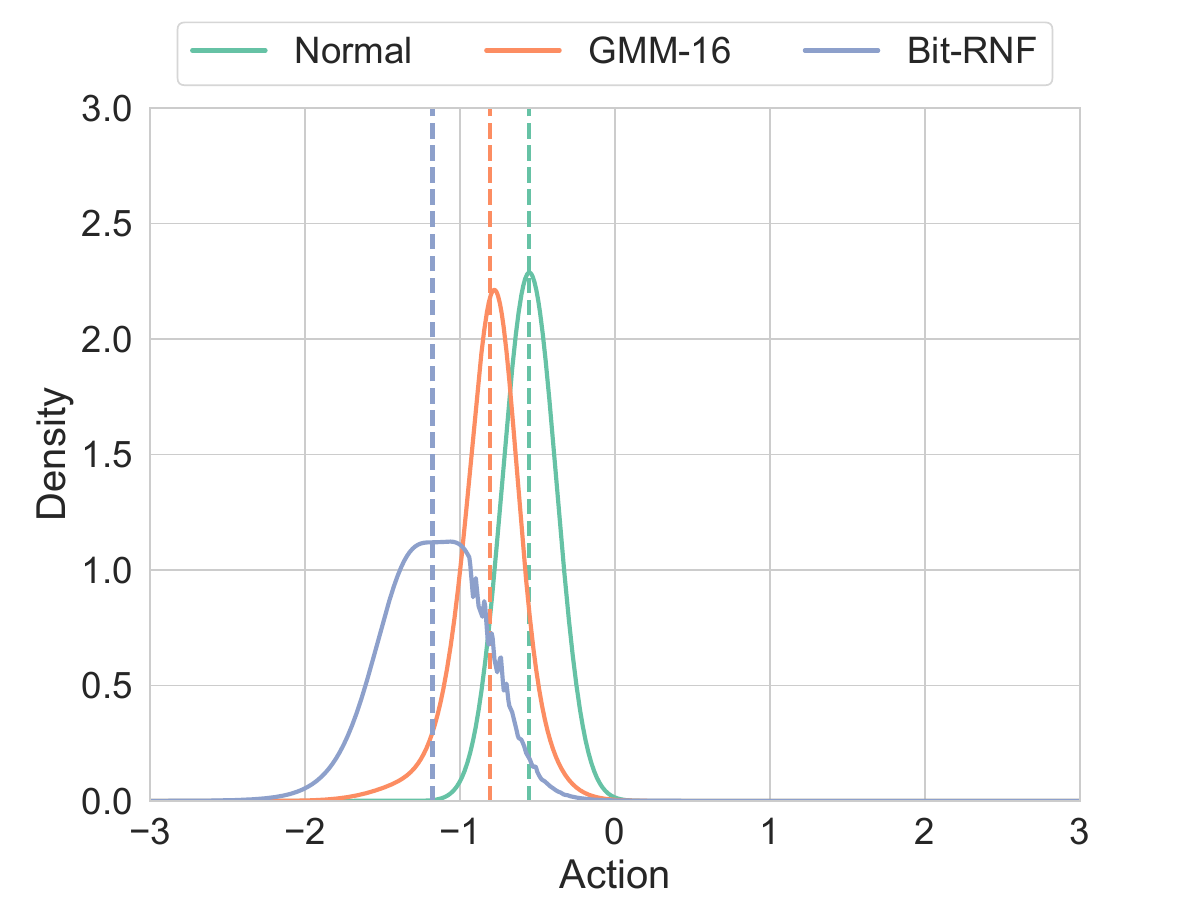}
        \subcaption{Flat shape}
        \label{fig:example_policy_flat}
    \end{subfigure}
    \caption{Examples of the trained policy shape:
        dashed lines indicate the means of the corresponding policies;
        GMM-16 converged to a shape not much different from Normal, while Bit-RNF acquired a variety of non-Gaussian distribution shapes.
    }
    \label{fig:example_policy}
\end{figure}

%Table
\begin{table}[tb]
    \caption{The sum of rewards after training:
        the mean and standard deviation for the 100 tests are listed;
        for each test, the analytic mean of the policy was used as action;
        successful cases are noted in bold.
    }
    \label{tab:sim_ablation}
    \centering
    \begin{tabular}{l cccc}
        \hline\hline
        Method & DoublePendulum & Hopper & HalfCheetah & Ant
        \\
        Bit-RNF
        & 8586 $\pm$ 2037
        & 2199 $\pm$ 583
        & 2119 $\pm$ 321
        & 2934 $\pm$ 236
        \\
        \hline
        RNF
        & N/A
        & N/A
        & N/A
        & N/A
        \\
        Bit
        & 7295 $\pm$ 2478
        & 1895 $\pm$ 922
        & \textbf{2010} $\pm$ 487
        & \textbf{3024} $\pm$ 265
        \\
        \hline\hline
    \end{tabular}
\end{table}

At first, we test numerical simulation tasks on Pybullet~\cite{brockman2016openai,coumans2016pybullet}, which is provided as benchmarks for RL.
Specifically, the following four tasks are trained with the five policy models described above:
\textit{InvertedDoublePendulumBulletEnv-v0} (DoublePendulum);
\textit{HopperBulletEnv-v0} (Hopper);
\textit{HalfCheetahBulletEnv-v0} (HalfCheetah);
and \textit{AntBulletEnv-v0} (Ant).
DNNs were trained 18 times under the same conditions with different random seeds, and the statistics were evaluated to compare the performance.
Note that another benchmark using SAC~\cite{haarnoja2018soft} was also shown in Appendix~\ref{app:benchmark}.

The sum of rewards at each episode is depicted in Fig.~\ref{fig:sim_score} as learning curve.
In addition, the 100 tests using the trained policy (more specifically, its mean as action) are summarized in Table~\ref{tab:sim_score}.
Note that \texttt{reward\_threshold} in this table, which is given in the official github repository~\cite{coumans2016pybullet}, represents the roughly maximum score, although the locomotion tasks (other than DoublePendulum) and DoublePendulum are successful with the score over 2000 and 8000, respectively.
In addition, the standard deviations are also listed in Table~\ref{tab:sim_score}, but it is difficult to correctly find significant differences based on the overlap of them because because the actual scores are distributed in a multimodal manner due to the success or failure of the task and the existence of multiple local solutions.

Comparing the learning curves, it can be seen that the performance of normal distribution was clearly lower than that of the other elaborated distribution models.
In student-t distribution, despite the increase of only one parameter $\nu$, the overall performance improved from that of normal distribution, which is consistent with the previous report~\cite{kobayashi2019student}.
GMM-10 generally achieved superior efficiency and score, but in HalfCheetah, it gained the lowest performance.
This may be because GMM-10 was prone to overlearn the local solution of pushing the body forward, and it was not able to fully use its abundant expressiveness.
On the other hand, GMM-16 succeeded in all the tasks.
However, the maximum computational cost was also observed compared to the other models (also see the next section), confirming that GMM-16 is not suitable for real-time control.

In contrast, Bit-RNF achieved the faster and successful learning in all the tasks.
This may be due to the fact that Bit-RNF has high expressiveness comparable to that of GMM-16 and succeeded in stable learning without falling into local solutions.
This stable learning would be obtained from that the conservative learning ability is inherited by using student-t distribution as the base, and as pointed out by the literature~\cite{behrmann2021understanding}, NF has implicit regularization to avoid excessive representation.
However, it should be noticed that Bit-RNF does not yield dramatic performance improvements over previous models, but is rather a way to fill in the last mile.

Here, we check the distribution shape of the acquired policies.
Since it is difficult to visualize them except for ones learned DoublePendulum, we collected state trajectories using Normal, GMM-16, and Bit-RNF for successful trials of DoublePendulum, and the distribution shapes of the three policies for the same state are illustrated in Fig.~\ref{fig:example_policy}.
It can be seen that Bit-RNF, as expected, changed to a more diverse distribution shape depending on the state, including asymmetric, multimodal, and small kurtosis shapes.
On the other hand, GMM-16 converged to a unimodal distribution, which may be due to the fact that DoublePendulum does not require the complex policy and, as mentioned above, mixture distributions are prone to overlearning in RL.

Finally, as ablation studies, we also tested RNF with normal distribution as its base and Bit without RNF but with bimodal student-t distribution.
Unfortunately, RNF has no record because numerical instability, causing NaN during training.
This can be ignored in the case where the base tail is heavy and extreme gradients are difficult to be generated, as in the proposed Bit-RNF (even with large $\tau$), but in normal distribution, $\tau$ must be small to avoid this.
The remaining Bit obtained the results shown in Table~\ref{tab:sim_ablation}.
While it outperformed Bit-RNF in Ant task, it was inferior in other tasks.
Thus, these ablation studies suggested that the proposed Bit-RNF certainly yields benefits from both RNF and design of the base distribution.

%%%%%%%%%%%%%%%%%%%%%%
\subsection{Demonstration}

%Figure
\begin{figure}[tb]
    \centering
    \includegraphics[keepaspectratio=true,width=0.96\linewidth]{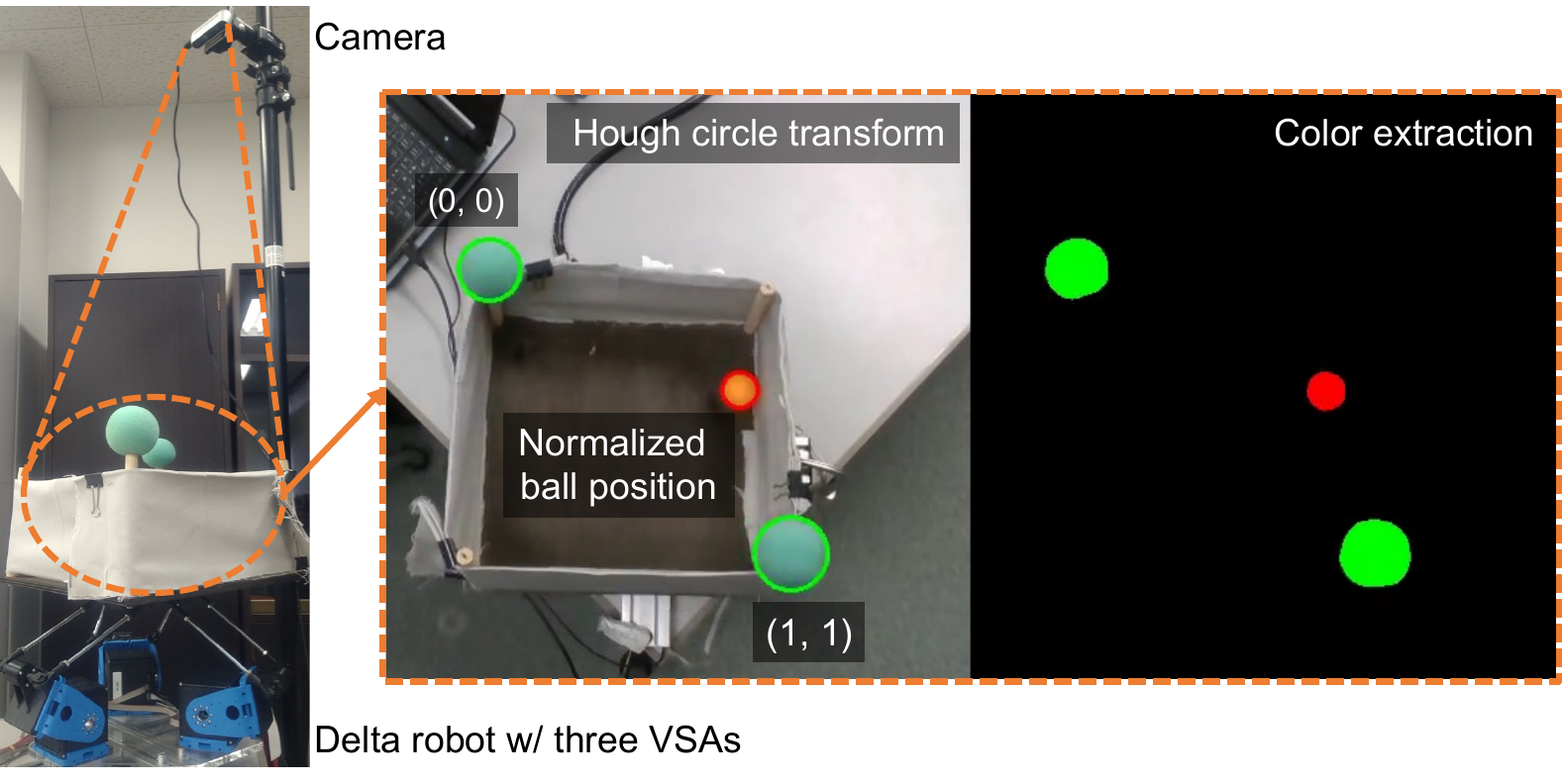}
    \caption{Ball-plate system on delta robot:
        by moving the base delta robot, the 0.3~m square plate fixed on its top and the ping-pong ball rolling on the plate can be moved;
        the ball position and movement are measured by a camera mounted overhead.
    }
    \label{fig:demo_env}
\end{figure}

%Figure
\begin{figure}[tb]
    \centering
    \includegraphics[keepaspectratio=true,width=0.96\linewidth]{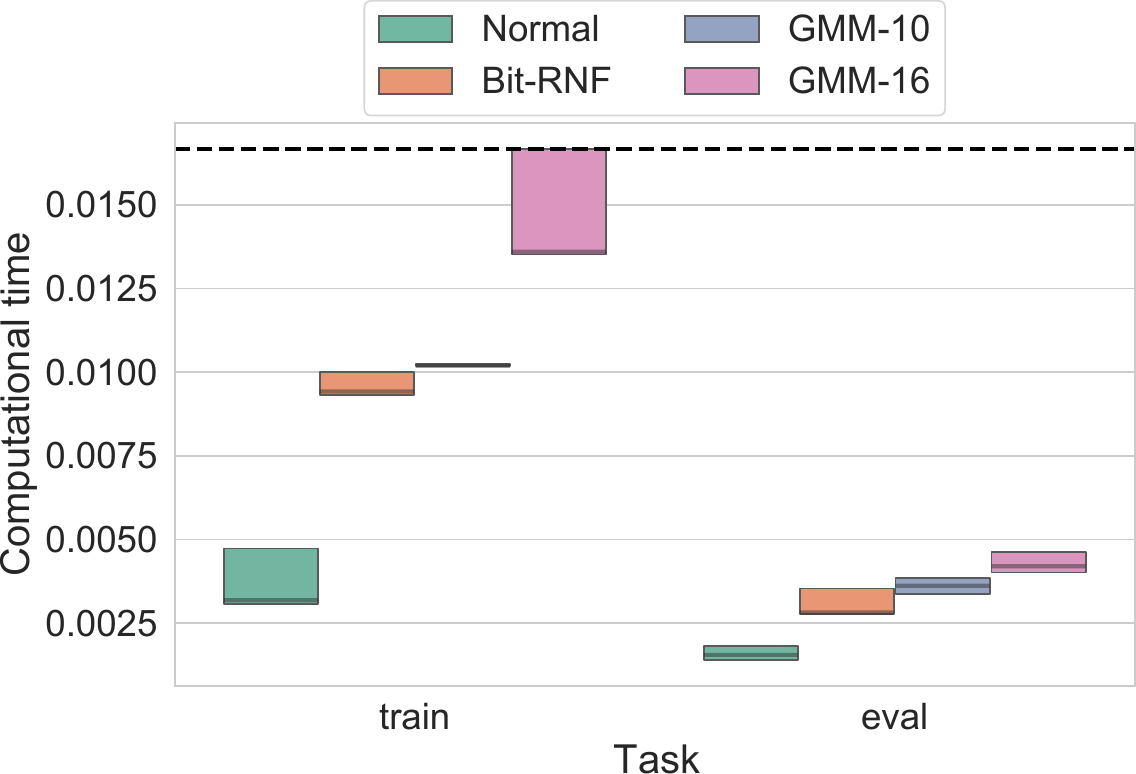}
    \caption{Worst-case computational time:
        ten episodes were conducted for each model, and the worst-case computational time for each episode was summarized;
        the dashed line denotes the desired period, and the worst case must be smaller than this;
        except GMM-16 during training, this requirement is satisfied.
    }
    \label{fig:demo_time}
\end{figure}

%Figure
\begin{figure}[tb]
    \begin{subfigure}[b]{0.48\linewidth}
        \centering
        \includegraphics[keepaspectratio=true,width=\linewidth]{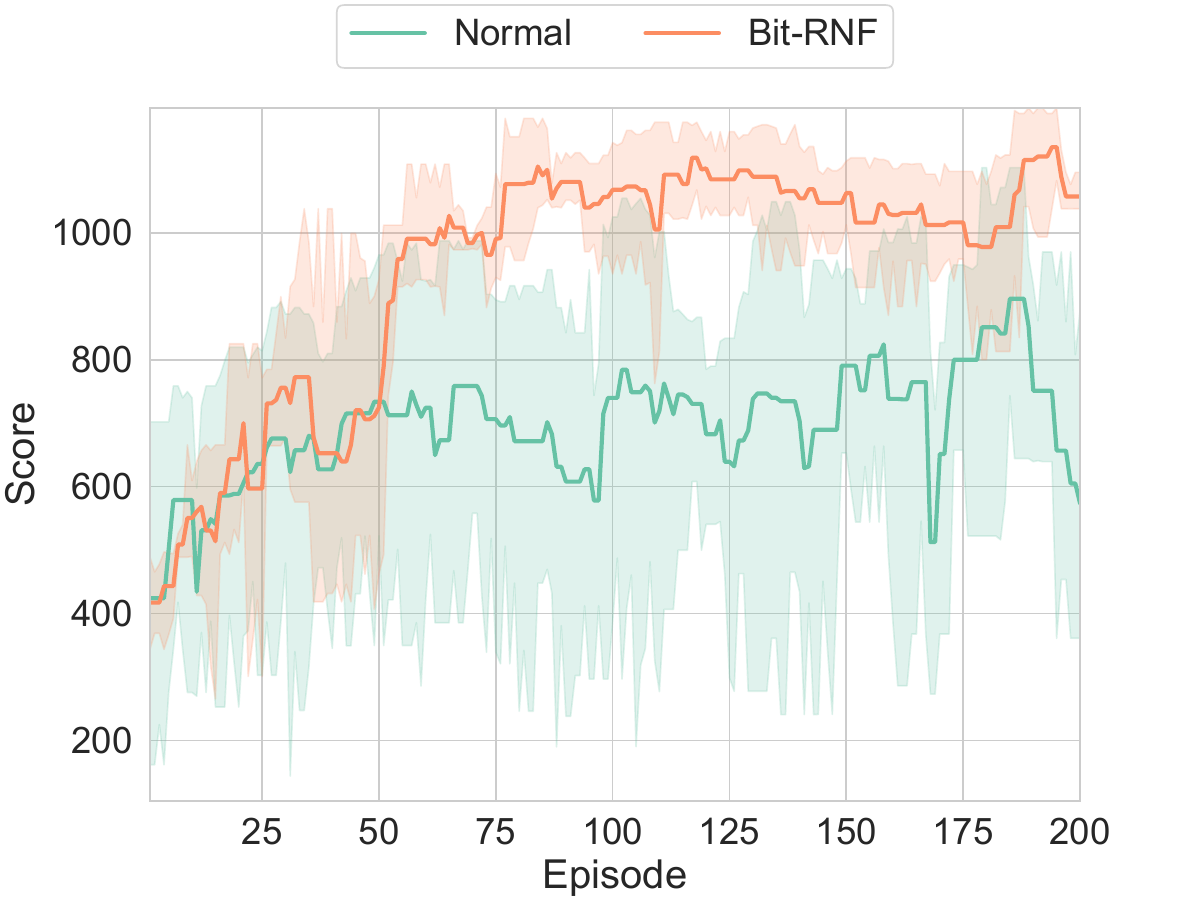}
        \subcaption{Learning curves}
        \label{fig:demo_score_learn}
    \end{subfigure}
    \begin{subfigure}[b]{0.48\linewidth}
        \centering
        \includegraphics[keepaspectratio=true,width=\linewidth]{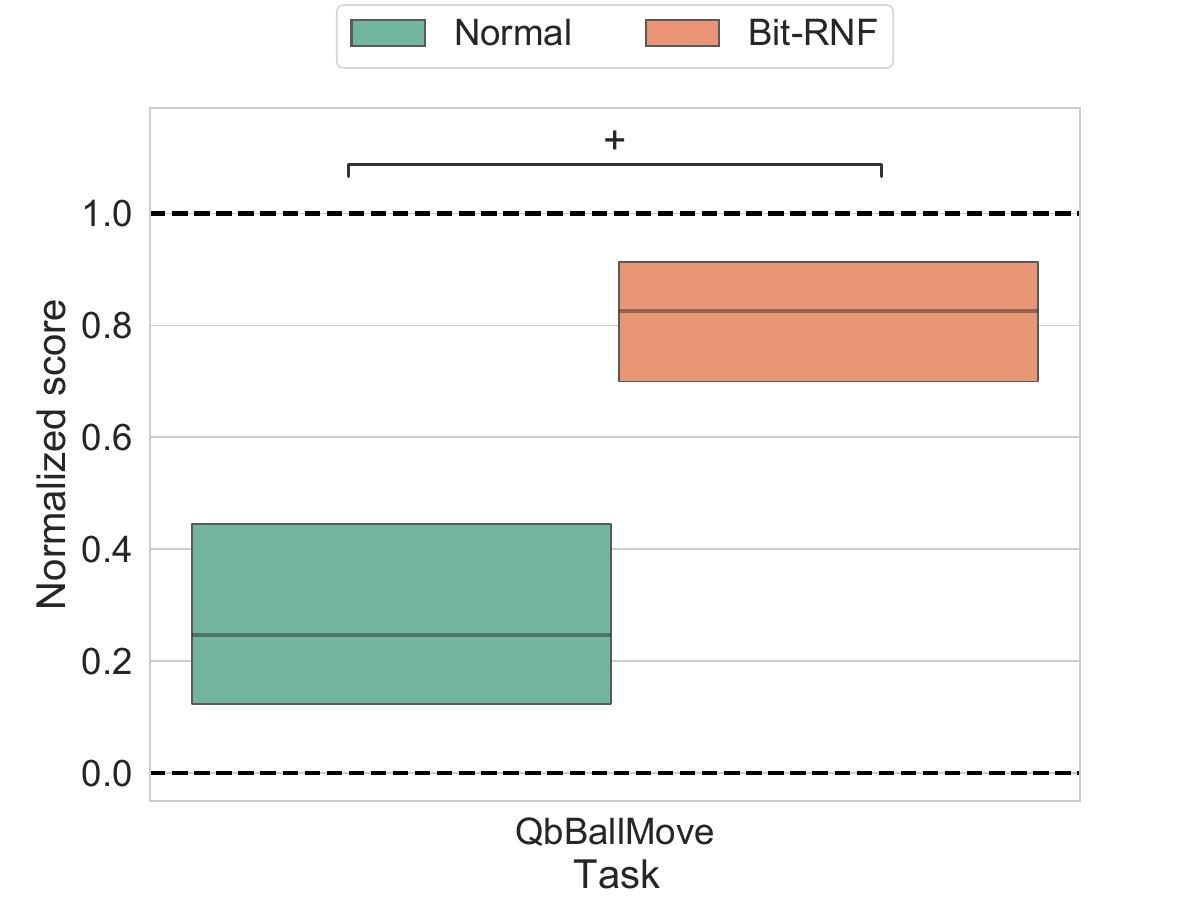}
        \subcaption{Test results}
        \label{fig:demo_score_test}
    \end{subfigure}
    \caption{Demonstration results:
        despite the small number of trials(i.e. three trials), we can confirm that Bit-RNF performed better than normal distribution, which is a common model, during and after learning.
    }
    \label{fig:demo_score}
\end{figure}

%Figure
\begin{figure}[tb]
    \begin{subfigure}[b]{0.48\linewidth}
        \centering
        \includegraphics[keepaspectratio=true,width=\linewidth]{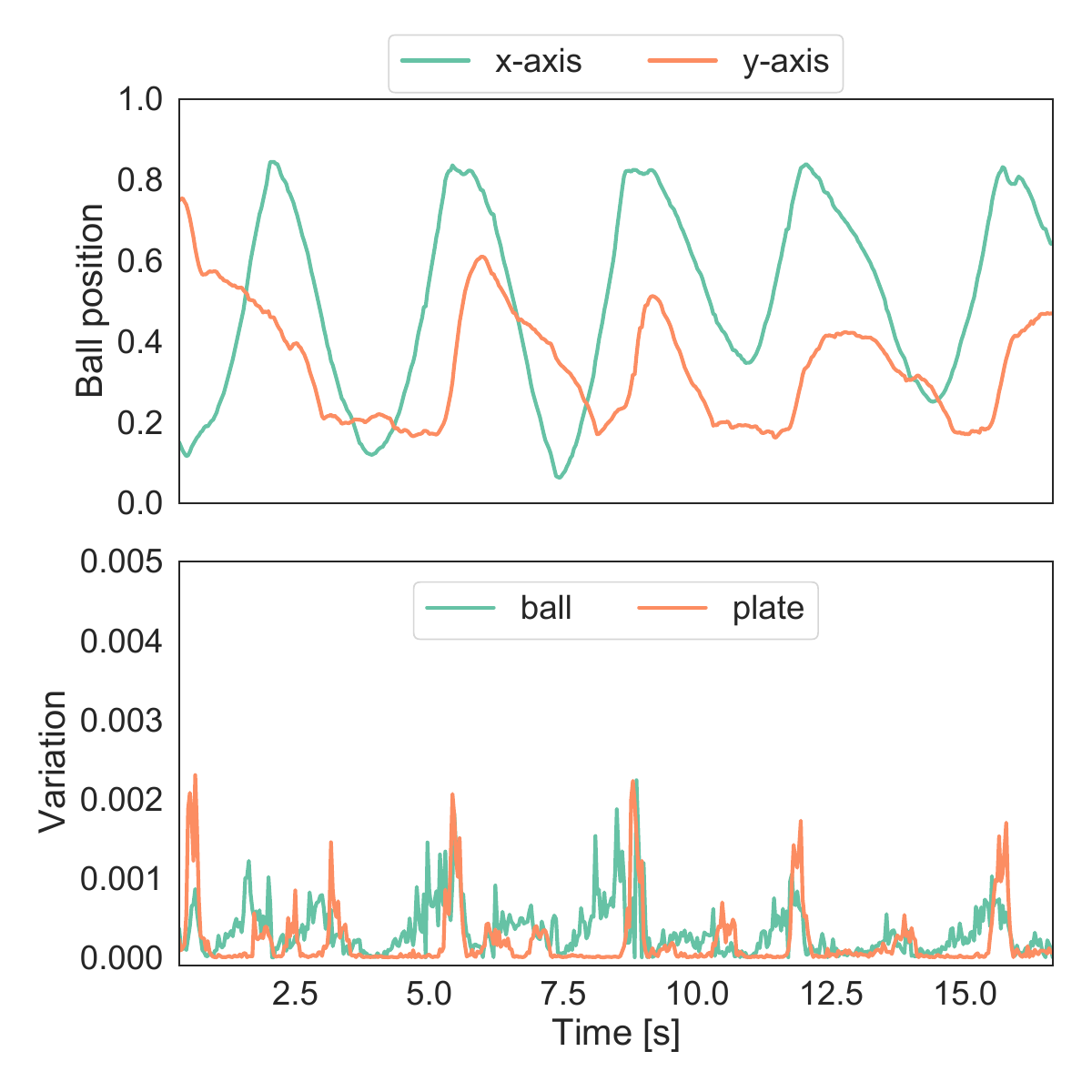}
        \subcaption{Normal}
        \label{fig:demo_traj_learn}
    \end{subfigure}
    \begin{subfigure}[b]{0.48\linewidth}
        \centering
        \includegraphics[keepaspectratio=true,width=\linewidth]{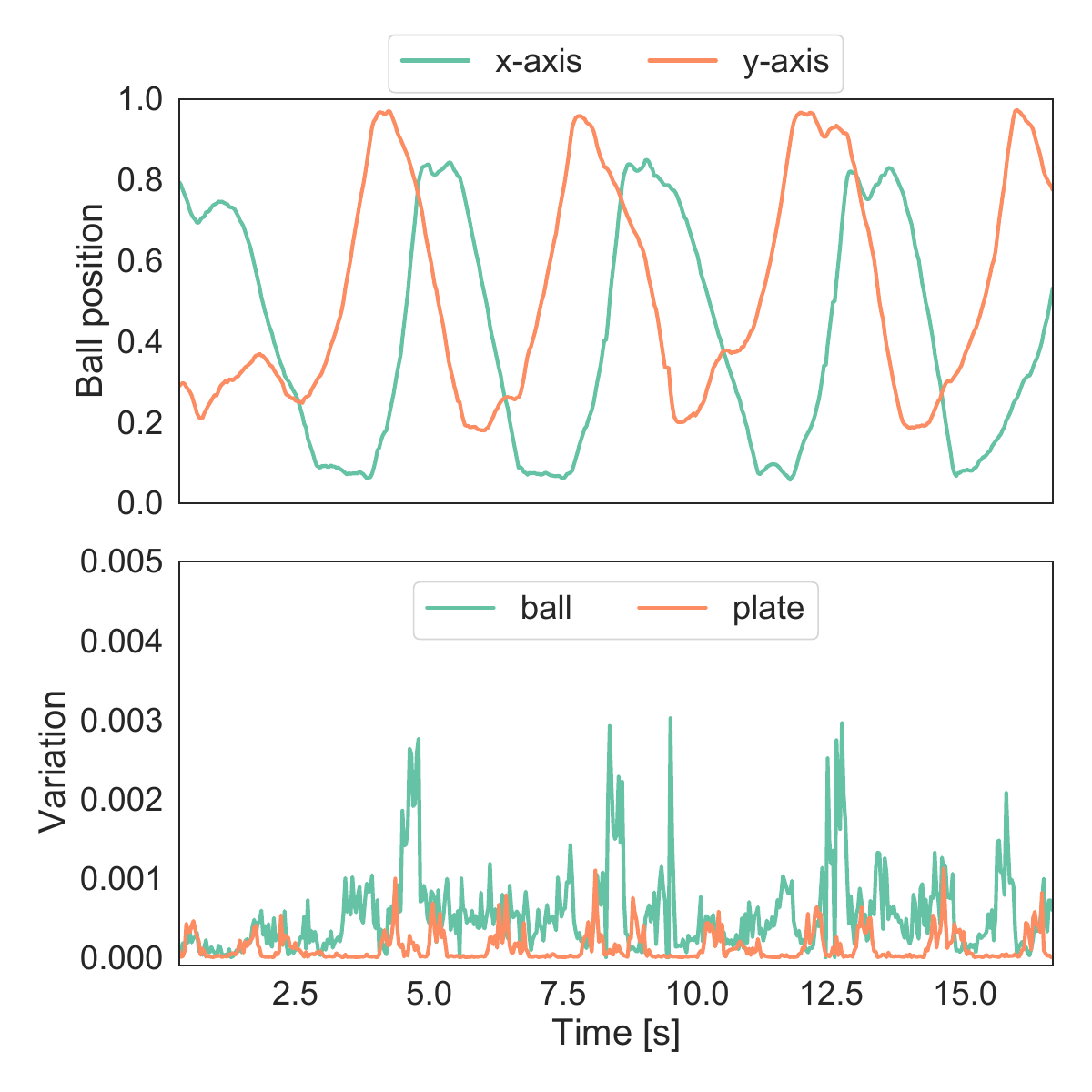}
        \subcaption{Bit-RNF}
        \label{fig:demo_traj_test}
    \end{subfigure}
    \caption{Demonstration trajectories:
        upper graph illustrates the $xy$-axes ball positions, showing that (a) Normal could move the ball mainly in $x$-axis, while (b) Bit-RNF could move the ball in a circular motion;
        lower graph shows the variations of ball and plate movements, indicating that (a) Normal periodically actuated the plate significantly but did not lead to the ball movement, while (b) Bit-RNF achieved the large ball movement with less plate movement.
    }
    \label{fig:demo_traj}
\end{figure}

To demonstrate applicability of Bit-RNF to a real robot control, a ball-plate system constructed by a delta robot (see Fig.~\ref{fig:demo_env}) is controlled.
The detailed system configuration is described in Appendix~\ref{app:demonstration}.
In summary, this task is to maximize the movement of the ball on the top of the plate by controlling the position (and stiffness) of the delta robot.

The control period of this system is 30~fps, but the waiting time between the agent action and the state transition of the environment is given to be half of that period.
Therefore, the actual control period during which the agent must decide its action is over 60~fps ($\simeq$ 0.016~sec).
The computational time during training (\textit{train}) and testing (\textit{eval}) was pre-evaluated using Intel Core i7-9750H, as shown in Fig.~\ref{fig:demo_time}.
During testing, all models were able to decide actions fast enough.
During training, however, the worst-case result of GMM-16 was over the desired period.
In contrast, Bit-RNF was stable enough and well within the desired period.
Note that GMM-10 could hold this requirement of the computational time, we omit it since the above simulations already indicated that its performance is inferior to that of Bit-RNF.

Based on the above results, two models that satisfy the desired computational period sufficiently, the representative normal distribution and the proposed Bit-RNF, are trained three times.
Note that the learning conditions are basically the same as in the simulation benchmarks (see Table~\ref{tab:param}), but the learning rate was reduced to $10^{-5}$ in order to mitigate the adverse effects of noise in the real environment.
The learning results are shown in Fig.~\ref{fig:demo_score}.
In addition, the behaviors generated by the trained policy are illustrated in the attached video.

As can be seen in the attached video, before learning, the ball could hardly be moved, or the robot was unable to respond after the ball was rolled.
by training, as shown in Fig.~\ref{fig:demo_score_learn}, only Bit-RNF consistently improved its score to a high level.
In addition, the post-training score (see Fig.~\ref{fig:demo_score_test}) indicated that Bit-RNF outperformed normal distribution.
The best results after learning are illustrated in Fig.~\ref{fig:demo_traj} and the attached video, which shows that Bit-RNF acquired energy-efficient motion by making the ball roll around on the plate, whereas normal distribution only moved the ball left and right, which is wasteful.
Indeed, we evaluated the efficiency of these motions as the sum of ball movement divided by the sum of plate movement.
As a result, we confirmed that Bit-RNF ($0.27/0.06 \simeq 4.3$) outperformed normal distribution ($0.16/0.1 \simeq 1.6$).
Thus, it is suggested that Bit-RNF contributes to task performance with higher expressiveness than the commonly used normal distribution, while being computationally cost-effective for practical use.

%%%%%%%%%%%%%%%%%%%%%%%%%%%%%%%%%%%%%%%%%%%%%%%%%%%%%%%%%%%%%%%%%%%%%%%%%%%%%%%%
\section{Conclusion and discussion}

%%%%%%%%%%%%%%%%%%%%%%
\subsection{Conclusion}

This paper derived a restricted normalizing flow (RNF) with the constraints that make it possible to compute the analytic mean of NF.
We mathematically showed that RNF can be achieved by i) separating the mean and the probability component of the symmetric probability distribution and ii) applying the invertible transformation by the odd function only to the probability component.
As examples, we introduced RNF versions of LRS flow and RealNVP, respectively.
In addition, the inexpressible distributions by the constraints, in particular, asymmetric and heavy-tailed distributions, were mostly taken back by setting bimodal student-t distribution as the base, so-called Bit-RNF.
The usefulness of the proposed model was statistically evaluated through numerical simulations, confirming improved learning efficiency and stability.
Finally, we demonstrated Bit-RNF on the ball-plate system constructed by the delta robot with VSAs.
Bit-RNF was able to adequately acquire this task, whereas the representative normal distribution was unsuccessful and GMM with 16 components, which has the comparable performance to Bit-RNF in the simulations, could not be computed within the control period.

%%%%%%%%%%%%%%%%%%%%%%
\subsection{Discussion for future work}

Although RNF proposed in this paper can be said to be with a sufficient condition for obtaining the analytic mean, it is not obvious that it is with a necessary condition.
Further analysis and design will make it possible to construct a new RNF that is with a necessary and sufficient condition, thereby holding the expressiveness of RNF as much as possible.
In addition, other statistics, such as mode, variance, and so on, would be more important for generating the optimal action and exploration.
Unfortunately, we are still unable to derive appropriate restrictions for obtaining their closed-form solutions.
Therefore, we further analyze NF to give the desired statistics in closed form.
Regarding the expressiveness, it is desirable to clarify what kind of problems require it.
For example, in problems with high-dimensional action space, RNF may be more likely to produce valuable exploration than simple noise by normal distribution.

One of the advantages of RNF would be its capability to represent the target distribution similar to the base.
This advantage can be used to add auxiliary objective(s) to the base, as mentioned in the introduction ~\cite{gambardella2019transflow,codevilla2018end,abdal2021styleflow}.
In RL, for example, the exploitation-exploration dilemma is widely known, and the shape (in particular, tail) of the policy is in charge of this exploration.
For this situation, by learning the base in the usual way to obtain the exploitation, and learning the target as the complementary distribution of the base (like~\cite{mukaeda2020development}), it should be possible to switch explicitly between the exploitation and the exploration depending on whether sampling from the base or the target.
Not only this example, but there are many real problems that deal with auxiliary objectives in addition to the main objective (such as entropy maximization~\cite{haarnoja2018soft}), and RNF can achieve constrained auxiliary objective optimization that does not compromise the main objective.

On the other hand, while this paper focuses on the use of RNF to model the policy in RL, it can be used to model a variety of probability distributions.
For example, in model-based RL~\cite{janner2019trust,aotani2021meta}, the state transition probability (and reward function) should be learned for planning the optimal action.
The state transition probability is also modeled as normal distribution in many cases, but their average behaviors would be correctly predicted by expressing it in RNF, while capturing the uncertainty more appropriately.
Indeed, prior studies have appropriately captured the complexity of state transitions by utilizing GMM~\cite{okada2020variational,okada2020planet} and non-restricted NF~\cite{power2022variational}.
Furthermore, this uncertainty may be ignored for fast planning of the optimal action, and in such cases, the analytic mean of RNF would be useful.

If the proposed method is utilized to extract a world model~\cite{okada2020planet,hafner2019learning}, for example, we would expect to extract low-dimensional latent variables without trade-off between expressiveness of observation and well-compressed continuous space, as like the literature~\cite{bhalodia2020dpvaes}.
In addition, although this paper concerned the fact that RNF is restricted to symmetric distributions, the latent variables symmetrically distributed would enable to make the latent variables, whose meaning is uncertain, more valuable and clearer.
Alternatively, robots that handle physical contacts have received increasing attention in recent years~\cite{modares2015optimized,kobayashi2021whole,itadera2021towards}, and discrete system changes by such physical contacts can be represented in the latent space by the base (without contacts) and the target (with contacts).

The complex distribution shape like NF is also useful for adversarial learning~\cite{mohaghegh2020advflow}.
Indeed, RL can increase robustness of the policy to more uncertain environments (e.g. partially observable ones~\cite{kurniawati2022partially}) by learning the policy that can resist a variety of attacks from an adversary~\cite{pinto2017robust}.
Since RNF can adjust the mean of attacks, it should promote robustness without introducing a bias from the nominal policy that are not attacked.

As described above, we would like to clarify the potential of RNF by utilizing it as a general-purpose stochastic model for various problems, rather than limiting it to a simple policy modeling method.

%%%%%%%%%%%%%%%%%%%%%%%%%%%%%%%%%%%%%%%%%%%%%%%%%%%%%%%%%%%%%%%%%%%%%%%%%%%%%%%%
\section*{Acknowledgement}

This work was supported by JST, PRESTO Grant Number JPMJPR20C3, Japan.

%%%%%%%%%%%%%%%%%%%%%%%%%%%%%%%%%%%%%%%%%%%%%%%%%%%%%%%%%%%%%%%%%%%%%%%%%%%%%%%%
\bibliographystyle{tfnlm}
\bibliography{biblio}

%%%%%%%%%%%%%%%%%%%%%%%%%%%%%%%%%%%%%%%%%%%%%%%%%%%%%%%%%%%%%%%%%%%%%%%%%%%%%%%%
\appendix

%%%%%%%%%%%%%%%%%%%%%%%%%%%%%%%%%%%%%%%%%%%%%%%%%%%%%%%%%%%%%%%%%%%%%%%%%%%%%%%%
\section{RealNVP}
\label{app:realnvp}

RealNVP~\cite{dinh2016density} is the neural-network-based transformations.
Although this paper does not employ RealNVP, we briefly introduce it as another type of NF (a.k.a. the affine transformation type), then, we clarified that it can be restricted as RNF.

When $x_b$ is $D > 1$-dimensional, $g: x_b \to x_t$ of RealNVP is given as follows:
\begin{align}
    x_t &= m \odot x_b + (1 - m) \odot \{ s(m \odot x_b; \theta_g)) \odot x_b + t(m \odot x_b; \theta_g) \}
    \\
    x_b &= m \odot x_t + (1 - m) \odot [ s(m \odot x_t; \theta_g))^{-1} \odot \{ x_t - t(m \odot x_t; \theta_g)) \} ]
\end{align}
where $m = \{0, 1\}^D$ denotes the random masking vector.
$s > \mathbb{R}_{+}$ and $t \in \mathbb{R}^D$ are scaling and translation functions constructed by the neural networks with $\theta_g$, respectively.
In that case, the determinant for Jacobian can be simply derived.
\begin{align}
    \left| \det \left( \frac{\partial g(x_b)}{\partial x_b}\right) \right| = m^\top s(m \odot x_b; \theta_g)
\end{align}

To obtain $g_\mathrm{odd}$ in RealNVP, $s$ and $t$ must be even and odd functions, respectively.
Such a restriction can be naively realized for $f(x; \theta): x \to y$ as below~\cite{mattheakis2019physical}.
\begin{align}
    y = \cfrac{1}{2} \left \{ f(x; \theta) \pm f(-x; \theta) \right \}
\end{align}
where $\pm$ symbol is positive/negative for the even/odd functions, respectively.

%%%%%%%%%%%%%%%%%%%%%%%%%%%%%%%%%%%%%%%%%%%%%%%%%%%%%%%%%%%%%%%%%%%%%%%%%%%%%%%%
\section{Details of nonlinear functions}
\label{app:nonlinear}

DNNs require activation function between layers and mapping function to make output within target domain (e.g. $[0, 1]$).
Such nonlinear functions are often given based on exponential function (e.g. sigmoid function), but their computational cost and stability are recently quentioned~\cite{barron2021squareplus}.
In this paper, we introduce implementations based on a ``squareplus'' function and its derivatives, which has been proposed in~\cite{barron2021squareplus}.
The features for obtained functions are two folds:
lightweight computational cost due to the non-use of exponential functions;
and stabilization of gradient calculations due to slow convergence to the edge of their domains.

Specifically the squareplus function has been defined as follows:
\begin{align}
    \mathrm{squareplus}(x) = \cfrac{1}{2} \left (x + \sqrt{x^2 + b} \right ) > 0
\end{align}
where $b \geq 0$ denotes the smoothness around $x=0$, and $b=4$ is the default value.

The derivative of the squareplus function has the similar shape to the sigmoid function, hence we define it as a squmoid function.
\begin{align}
    \mathrm{squmoid}(x) = \cfrac{1}{2} \left ( \cfrac{x}{\sqrt{x^2+b}} + 1 \right ) \in (0, 1)
\end{align}

Noting that tanh function can be defined using the sigmoid function, we define the following squaresign function.
\begin{align}
    \mathrm{squaresign}(x) = 2 \mathrm{squmoid}(2x) - 1 = \cfrac{2x}{\sqrt{4x^2 + b}} \in (-1, 1)
\end{align}

In addition, the author of~\cite{barron2021squareplus} suggested squish function in his twitter discussion, which replaces the sigmoid function in swish function~\cite{elfwing2018sigmoid}, popular activation function, with the squmoid function.
\begin{align}
    \mathrm{squish}(x) = \mathrm{squmoid}(x) x = \cfrac{1}{2} \left ( \cfrac{x^2}{\sqrt{x^2+b}} + x \right )
\end{align}

Finally, softmax function to satisfy $\boldsymbol{1}^\top \mathrm{softmax}(x) = 1$ (with $\boldsymbol{1}$ the vector of ones) has also been reported as a bottleneck of representational capability due to linearity of its logarithm etc~\cite{kanai2018sigsoftmax}.
Therefore, we define squaremax function by replacing the exponential function in the softmax function with the squareplus function as below.
\begin{align}
    \mathrm{squaremax}(x) = \cfrac{\mathrm{squareplus}(x)}{\boldsymbol{1}^\top \mathrm{squareplus}(x)}
\end{align}

%%%%%%%%%%%%%%%%%%%%%%%%%%%%%%%%%%%%%%%%%%%%%%%%%%%%%%%%%%%%%%%%%%%%%%%%%%%%%%%%
\section{Details of distributions}
\label{app:distribution}

The implementation methods of the $D$-dimensional probability distribution models employed in the experiments are summarized here.
Since the parameters of each model have own domains, the real output from DNNs should be transformed by one of the above nonlinear functions.
Therefore, we describe the parameters of each model and their corresponding nonlinear functions.

First, diagonal normal distribution has the location parameter $\mu \in \mathbb{R}^D$ and the scale parameter $\sigma \in \mathbb{R}^D_{+}$.
Since $\mu$ can exist in the whole real space, nonlinear transformation is unnecessary.
On the other hand, $\sigma$ is restricted to positive real space, so it should be transformed using the squareplus function.

Second, diagonal student-t distribution has the degrees of freedom parameter $\nu \in \mathbb{R}_{+}$ in addition to $\mu$ and $\sigma$.
Here, we would like to consider designing $\nu$ according to $D$, as the literature~\cite{ilboudo2022adaterm}, so that its behavior is stable and adjustable regardless of the dimensionality of the action space.
To this end, from the correspondence with q-Gaussian, we obtain $\nu$ by the following equation.
\begin{align}
    \nu = \cfrac{2}{q - 1} - D
\end{align}
where $0 < q - 1 < 2/D$ for non-compact distribution.
Note, however, that $q - 1 < 1/D < 2/D$ must be satisfied to obtain the analytic mean with $\nu > 1$.
Then, $q - 1 \in (0, 1/D)$ is obtained using the squmoid function divided by $D$ and converted to $\nu$ in the above equation.

GMM has the mixture ratios for the respective components $\rho$, the sum of which must be $1$.
Therefore, $\rho$ for GMM can be computed using the squaremax function.

On the other hand, the bimodal distribution does not need $\rho$ for all the components since $\rho \in (0, 1)$ for one component is given, the other can be obtained by $1 - \rho$, as already described in eq.~\eqref{eq:bimodal}.
Therefore, $\rho$ for the bimodal distribution can be computed using the squmoid function.

%%%%%%%%%%%%%%%%%%%%%%%%%%%%%%%%%%%%%%%%%%%%%%%%%%%%%%%%%%%%%%%%%%%%%%%%%%%%%%%%
\section{Additional benchmark}
\label{app:benchmark}

%Figure
\begin{figure}[tb]
    \begin{subfigure}[b]{0.48\linewidth}
        \centering
        \includegraphics[keepaspectratio=true,width=\linewidth]{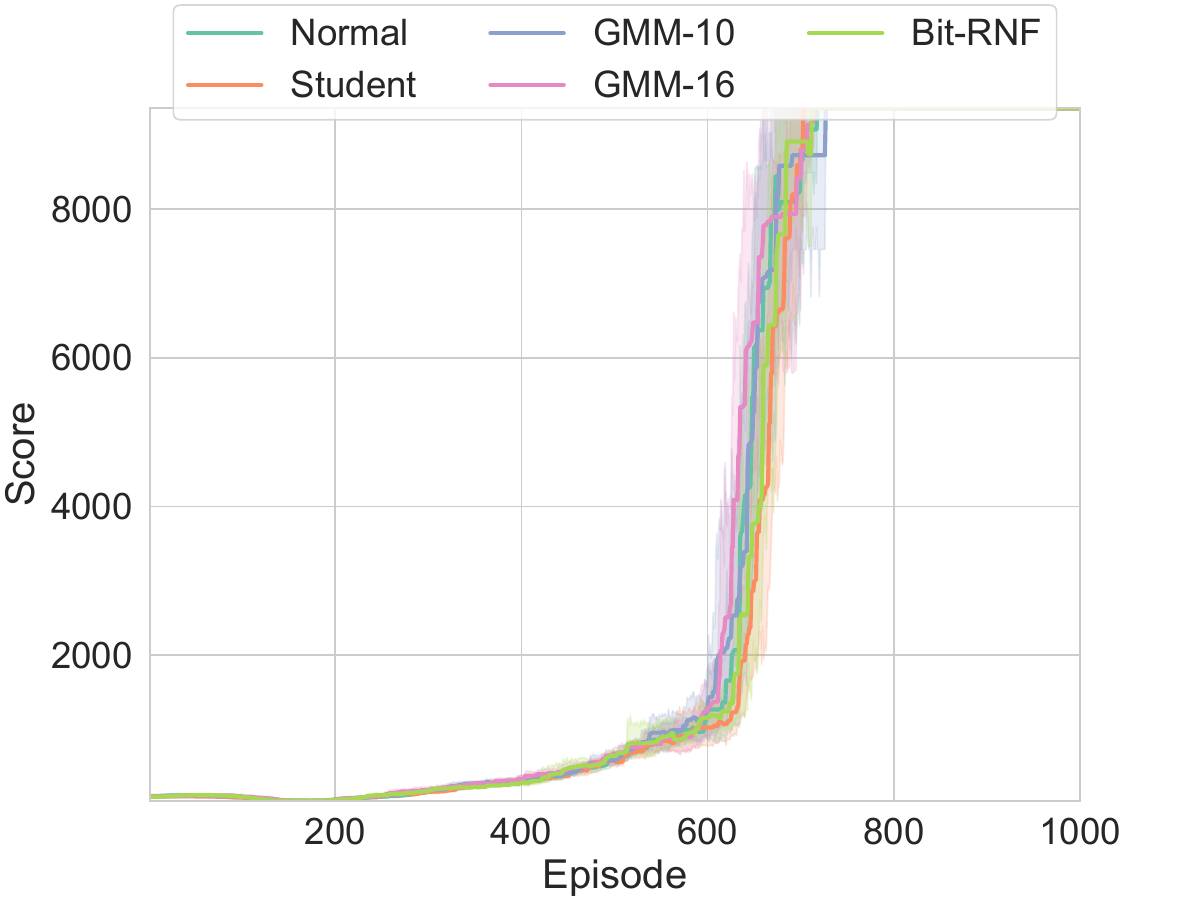}
        \subcaption{DoublePendulum}
        \label{fig:sac_score_DoublePendulum}
    \end{subfigure}
    \begin{subfigure}[b]{0.48\linewidth}
        \centering
        \includegraphics[keepaspectratio=true,width=\linewidth]{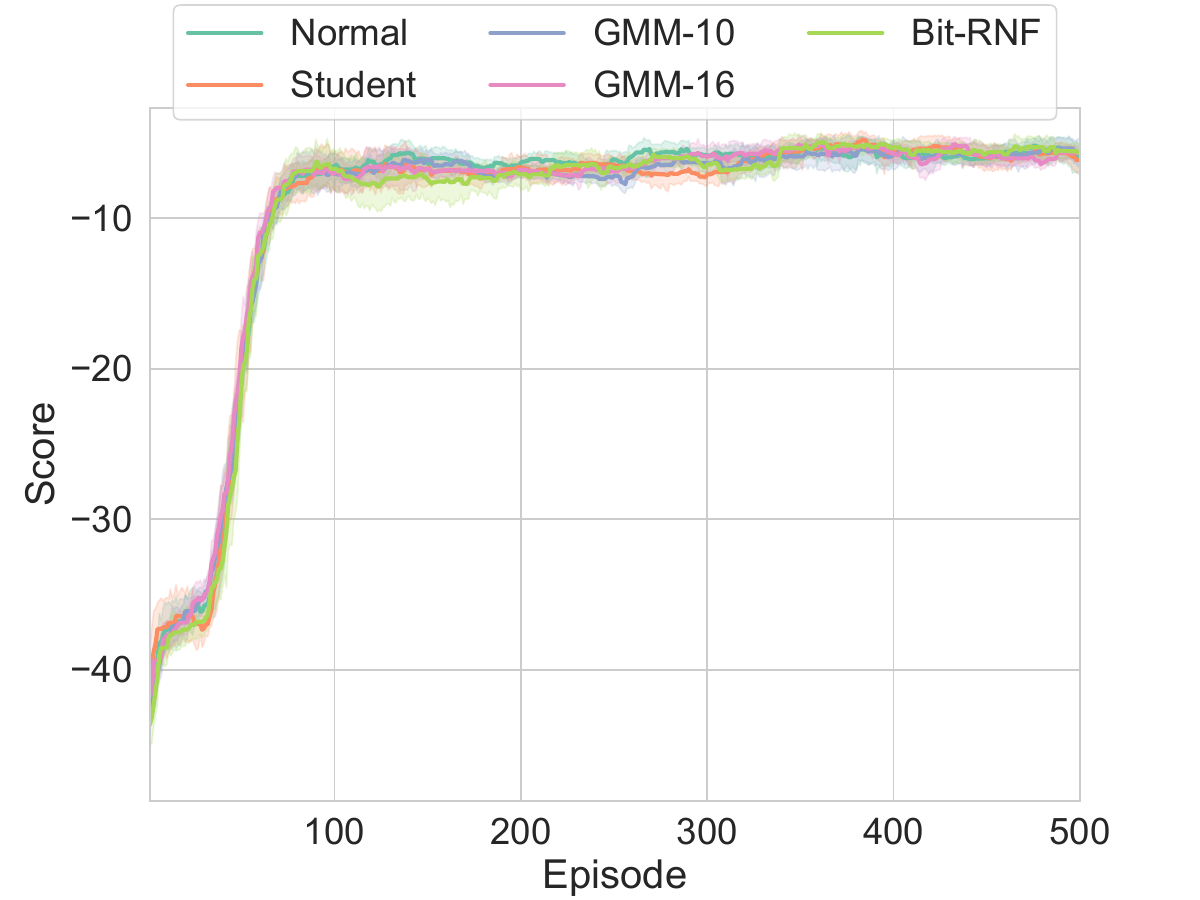}
        \subcaption{Reacher}
        \label{fig:sac_score_Hopper}
    \end{subfigure}
    \begin{subfigure}[b]{0.48\linewidth}
        \centering
        \includegraphics[keepaspectratio=true,width=\linewidth]{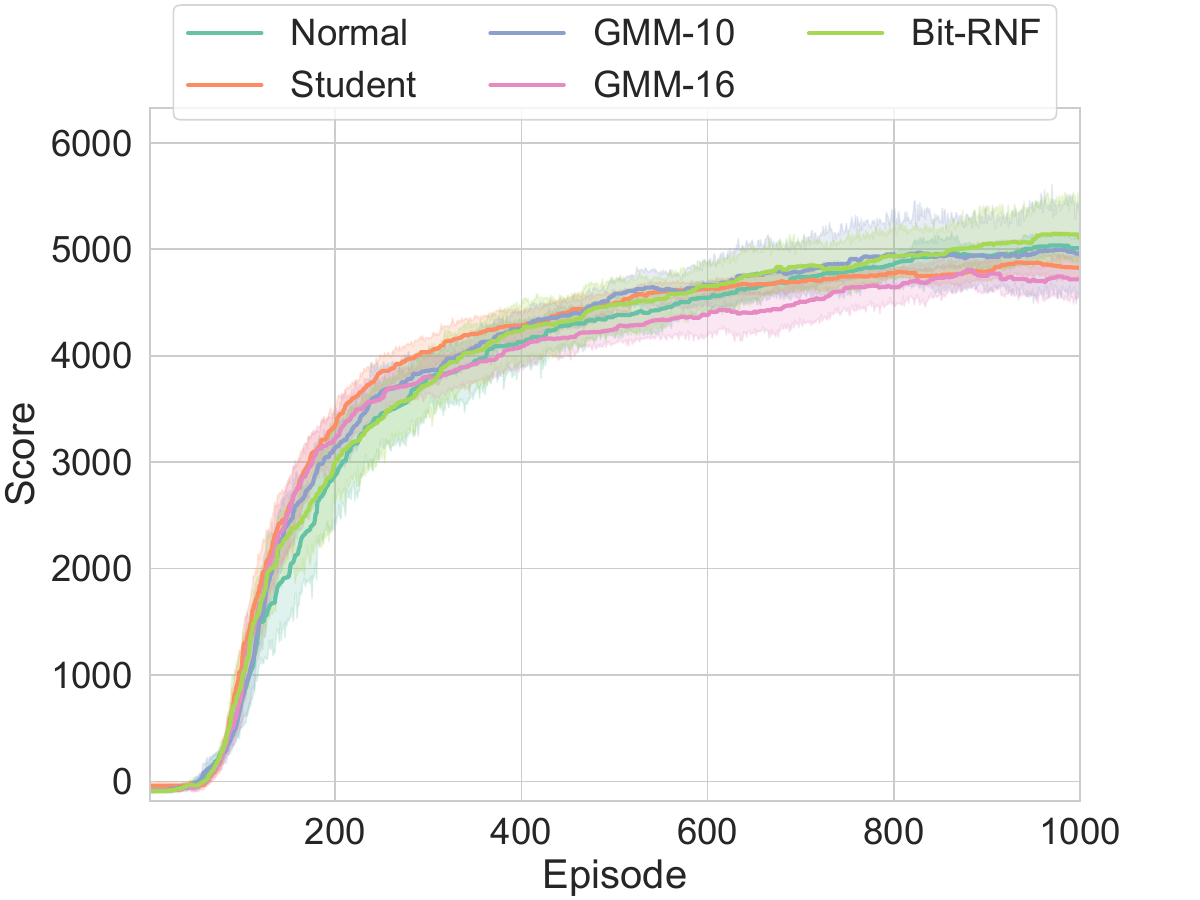}
        \subcaption{HalfCheetah}
        \label{fig:sac_score_HalfCheetah}
    \end{subfigure}
    \begin{subfigure}[b]{0.48\linewidth}
        \centering
        \includegraphics[keepaspectratio=true,width=\linewidth]{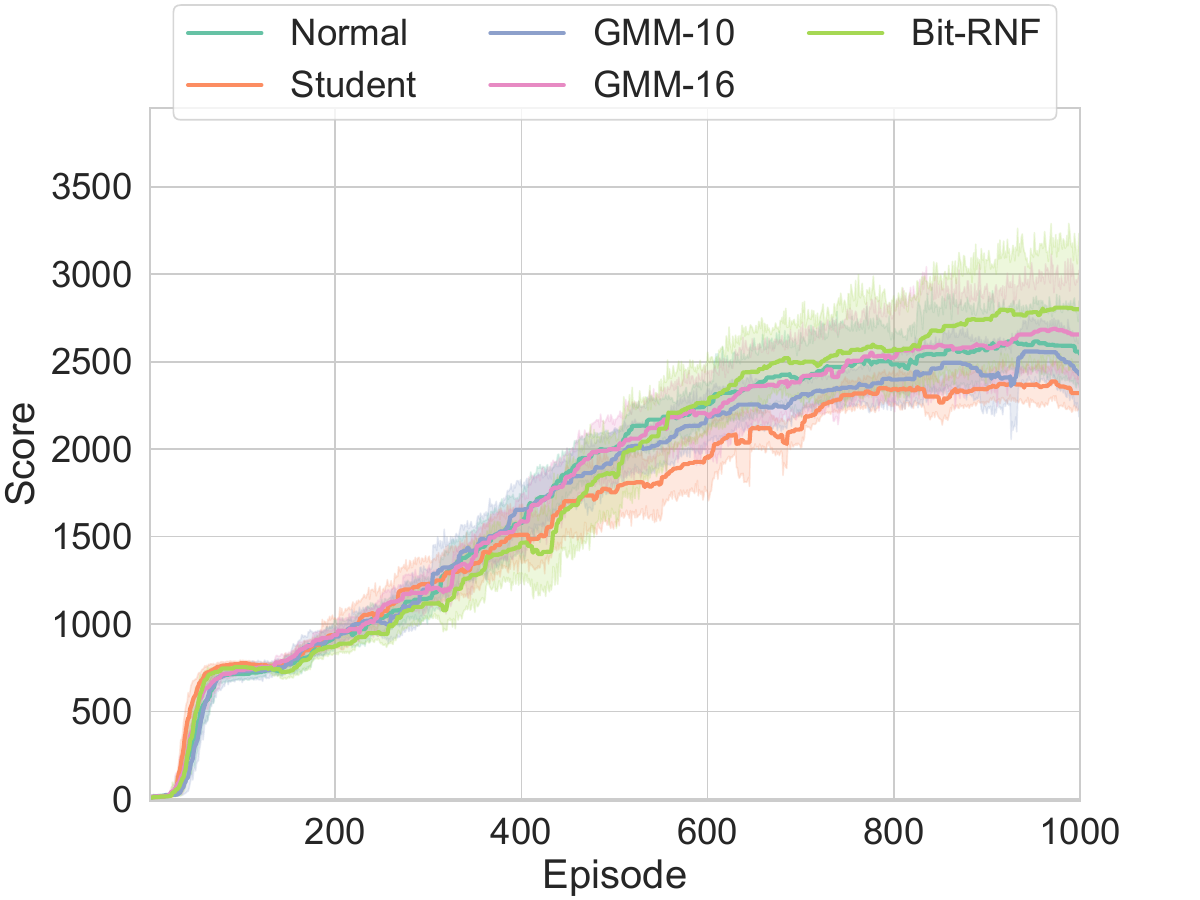}
        \subcaption{Ant}
        \label{fig:sac_score_Ant}
    \end{subfigure}
    \caption{Learning curves for the respective benchmark tasks:
        foo
    }
    \label{fig:sac_score}
\end{figure}
%Table
\begin{table}[tb]
    \caption{The sum of rewards after training:
        the mean and standard deviation for the 100 tests are listed;
        for each test, the analytic mean of the policy was used as action.
    }
    \label{tab:sac_score}
    \centering
    \begin{tabular}{l cccc}
        \hline\hline
        Method & DoublePendulum & Reacher & HalfCheetah & Ant
        \\
        \hline
        Normal
        & 9354 $\pm$ 9
        & -7.8 $\pm$ 0.7
        & 4820 $\pm$ 200
        & 1443 $\pm$ 437
        \\
        Student
        & 9352 $\pm$ 8
        & -8.8 $\pm$ 0.9
        & 4717 $\pm$ 195
        & 1222 $\pm$ 339
        \\
        GMM-10
        & 9345 $\pm$ 9
        & -8.7 $\pm$ 1.2
        & 4756 $\pm$ 817
        & 1219 $\pm$ 588
        \\
        GMM-16
        & 9348 $\pm$ 11
        & -8.9 $\pm$ 1.1
        & 4278 $\pm$ 537
        & 1799 $\pm$ 580
        \\
        Bit-RNF
        & 9348 $\pm$ 10
        & -8.3 $\pm$ 1.0
        & 4843 $\pm$ 363
        & 1770 $\pm$ 407
        \\
        \hline\hline
    \end{tabular}
\end{table}

We compare the performance of policy models with SAC~\cite{haarnoja2018soft}, a representative recent RL algorithm.
The implementation of SAC is based on the original paper and implementation, and the policy model used in the original algorithm (i.e. normal distribution) can be replaced to the ones introduced in the above simulations.
Note that auto-tuning a temperature parameter $\alpha$ is not used since its original implementation is regarded as an equality constraint even though it was claimed from an inequality constraint, namely improper implementation.
Instead, we set $\alpha=0.05$ empirically.

The tasks to be solved are employed from Mujoco~\cite{todorov2012mujoco}: \textit{InvertedDoublePendulum-v4} (DoublePendulum); \textit{Reacher-v4} (Reacher); \textit{HalfCheetah-v4} (HalfCheetah); and \textit{Ant-v4} (Ant).
Each task is initialized and trained 10 times with different random seeds for each method, and the performance of the learning curves and the polices after learning are compared.

The learning results are shown in Fig.~\ref{fig:sac_score}.
In addition, the 100 tests using the trained policy are summarized in Table~\ref{tab:sac_score}.
Both results show that there was little difference in performance between the policy models for DoublePendulum and Reacher, which have small action dimensions.
This would be due to the fact that entropy maximization by SAC facilitated the exploration sufficiently, and complex exploration was unnecessary.
On the other hand, there was a difference in learning results between HalfCheetah and Ant, where the action dimensions were relatively large, and only Bit-RNF achieved a higher learning curve and score for both tasks.
This may be because the expressiveness of the policy model was important in capturing the various behaviors generated by the combinations of explorations for each action dimension.

As a remark, the overall Student tends to be lower than Normal, which gives the impression of being different from the results in Section~\ref{subsec:benchmark}, but this is expected that the invertible transformation of the distribution by the tanh function, assuming that the action space is bounded in the SAC implementation.
In other words, the heavy tail in Student did not contribute much to the exploration, and the likelihood tends to be high near the boundary, which is suppressed by the entropy maximization of SAC, making it difficult to generate actions near the boundary.
In addition, as a more simple reason, the original SAC is implemented with normal distribution, and it should be well tuned for that model.

%%%%%%%%%%%%%%%%%%%%%%%%%%%%%%%%%%%%%%%%%%%%%%%%%%%%%%%%%%%%%%%%%%%%%%%%%%%%%%%%
\section{Details of demonstration}
\label{app:demonstration}

The robot system used for the demonstration, as shown in Fig.~\ref{fig:demo_env}, is with a qbmove Advanced Kit Delta developed by qbrobotics as a base.
Each actuator of this robot is a modular VSA~\cite{catalano2011vsa}, whose angle and stiffness can be controlled.
By solving its inverse kinematics analytically, the 3D tip position can be given as command.
In addition, the tip position can be estimated according to its forward kinematics.
Even though, since it is difficult to analyze the effects of variable stiffness on the system, RL control would be suitable for this system, as in the previous study~\cite{kobayashi2022optimization}.

This robot is generally suspended on the top, but in this demonstration, it is inverted and used as the base.
However, the weight of its gripper makes the linkage mechanism fall into singular postures.
To avoid this problem, the gripper was removed to reduce weight.
Instead, a 0.3~m square plate is directly fixed on the tip of the delta robot.

The plate is surrounded by a fabric, the thickness and tension of which vary in places, making the elastic modulus unknown.
When enclosing the fabric, four pillars are attached on all the corners, and two blue-green markers are placed on two of the them at opposite corners.
A ping-pong (orange) ball is placed on this plate and allowed to roll freely.

The ball and the two markers are detected by a camera (Intel RealSense D435i) mounted on the top.
Specifically, orange and blue-green colors are extracted from the RGB image acquired by the camera, respectively, to generate the respective binary images.
Afterwards, for each binary image, one (or two for the markers) circle is extracted by Hough transform, and the coordinates of the resulting circle are obtained.
The coordinates are normalized so that the marker on the upper left of the screen is at the origin $(0, 0)$ and the one on the lower right is at $(1, 1)$.
The ball position is measured in the normalized coordinates.

From this system, a total of 31 dimensions are measured or estimated as the environmental state, which is given to the RL agent.
Specifically, it contains the 3D tip position and deviation, the 2D ball position and deviation, the commanded values, and the dynamical features of each VSA (i.e. equilibrium point, stiffness, torque, and elastic energy).
The agent action space is four-dimensional, consisting of the 3D acceleration of the tip and the common stiffness for all the actuators.
After sending the agent action to the system as the command, the next state is obtained after waiting $1/30/2 \simeq 0.016$~sec.
In addition, the reward for the ball moving task is computed by the following formula.
\begin{align}
    r = \begin{cases}
        -1 & \mathtt{ball\_lost}(s) \land \mathtt{ball\_lost}(s^\prime)
        \\
        100 \|p_\mathrm{ball}^\prime - p_\mathrm{ball}\|_2^2 & \mathrm{otherwise}
    \end{cases}
\end{align}
where $p_\mathrm{ball}$ denotes the normalized ball position.

When the condition that the reward is $-1$ (i.e. the ball is lost) is satisfied, the episode is forcibly terminated.
The above one step is taken at 30~fps, and 1000 steps at maximum are performed during one episode.

%%%%%%%%%%%%%%%%%%%%%%%%%%%%%%%%%%%%%%%%%%%%%%%%%%%%%%%%%%%%%%%%%%%%%%%%%%%%%%%%
\end{document}